\documentclass{article}

\usepackage[final]{neurips_2022}

\usepackage[utf8]{inputenc} % allow utf-8 input
\usepackage[T1]{fontenc}    % use 8-bit T1 fonts
\usepackage{hyperref}       % hyperlinks
\usepackage{url}            % simple URL typesetting
\usepackage{booktabs}       % professional-quality tables
\usepackage{amsfonts}       % blackboard math symbols
\usepackage{nicefrac}       % compact symbols for 1/2, etc.
\usepackage{microtype}      % microtypography
\usepackage{xcolor}         % colors

\usepackage{paralist}
\usepackage{graphicx}
\usepackage{amsmath}
\usepackage{amsfonts}
\usepackage{multirow}
\usepackage{float}
\usepackage{xcolor}
\usepackage{wrapfig}
\usepackage{algorithmicx}
\usepackage[ruled]{algorithm}
\usepackage{algpseudocode}
\usepackage{arydshln}
\usepackage{caption}
\usepackage{subcaption}
\usepackage{adjustbox}
\usepackage{pythonhighlight}

\definecolor{darkblue}{rgb}{0.0, 0.0, 0.55} %darkblue
\definecolor{denim}{rgb}{0.08, 0.38, 0.74} %denim
\makeatletter
\def\mathcolor#1#{\@mathcolor{#1}}
\def\@mathcolor#1#2#3{%
  \protect\leavevmode
  \begingroup
    \color#1{#2}#3%
  \endgroup
}
\makeatother

\title{Contextual Squeeze-and-Excitation for Efficient\\
Few-Shot Image Classification}

\author{%
  Massimiliano Patacchiola\\
  University of Cambridge\\
  \texttt{mp2008@cam.ac.uk} \\
   \And
   John Bronskill\\
   University of Cambridge\\
   \texttt{jfb54@cam.ac.uk}\\
   \And
   Aliaksandra Shysheya\\
   University of Cambridge\\
   \texttt{as2975@cam.ac.uk}\\
   \And
   Katja Hofmann\\
   Microsoft Research\\
   \texttt{kahofman@microsoft.com} \\
   \And
   Sebastian Nowozin\thanks{Work done while the author was at Microsoft Research -- Cambridge (UK)}\\
   \texttt{nowozin@gmail.com} \\
    \\
   \And
   Richard E. Turner\\
   University of Cambridge\\
   \texttt{ret26@cam.ac.uk}\\
}

\begin{document}

\maketitle

\begin{abstract}
  Recent years have seen a growth in user-centric applications that require effective knowledge transfer across tasks in the low-data regime. An example is personalization, where a pretrained system is adapted by learning on small amounts of labeled data belonging to a specific user. This setting requires high accuracy under low computational complexity, therefore the Pareto frontier of accuracy vs.~adaptation cost plays a crucial role. In this paper we push this Pareto frontier in the few-shot image classification setting with a key contribution: a new adaptive block called Contextual Squeeze-and-Excitation (CaSE) that adjusts a pretrained neural network on a new task to significantly improve performance with a single forward pass of the user data (context). We use meta-trained CaSE blocks to conditionally adapt the body of a network and a fine-tuning routine to adapt a linear head, defining a method called UpperCaSE. UpperCaSE achieves a new state-of-the-art accuracy relative to meta-learners on the 26 datasets of VTAB+MD and on a challenging real-world personalization benchmark (ORBIT), narrowing the gap with leading fine-tuning methods with the benefit of orders of magnitude lower adaptation cost.
\end{abstract}

\section{Introduction}

In recent years, the growth of industrial applications based on recommendation systems \citep{bennett2007netflix}, speech recognition \citep{xiong2018microsoft}, and personalization \citep{massiceti2021orbit} has sparked an interest in machine learning techniques that are able to adapt a model on small amounts of data belonging to a specific user. 
A key factor in many of these applications is the Pareto frontier of accuracy vs.~computational complexity (cost to adapt). For example, in a real-time classification task on a phone, a pretrained model must be personalized by exploiting small amounts of data on the user's device (context). In these applications the goal is twofold: maximize the classification accuracy on unseen data (target) while avoiding any latency and excessive use of computational resources.

Methods developed to face these challenges in the few-shot classification setting can be grouped in two categories: meta-learning and fine-tuning. Meta-learning is based on the idea of learning-how-to-learn by improving the algorithm itself \citep{schmidhuber1987evolutionary, hospedales2020meta}. Meta-learners are trained across multiple tasks to ingest a labeled context set, adapt the model, and predict the class membership of an unlabeled target point. Fine-tuning methods adjust the parameters of a pretrained neural network on the task at hand by iterative gradient-updates \citep{chen2019closer, triantafillou2019meta, tian2020rethinking, kolesnikov2020big, dumoulin2021comparing}.

\begin{wrapfigure}{r}{0.4\textwidth}
  \centering
  \includegraphics[width=0.4 \textwidth]{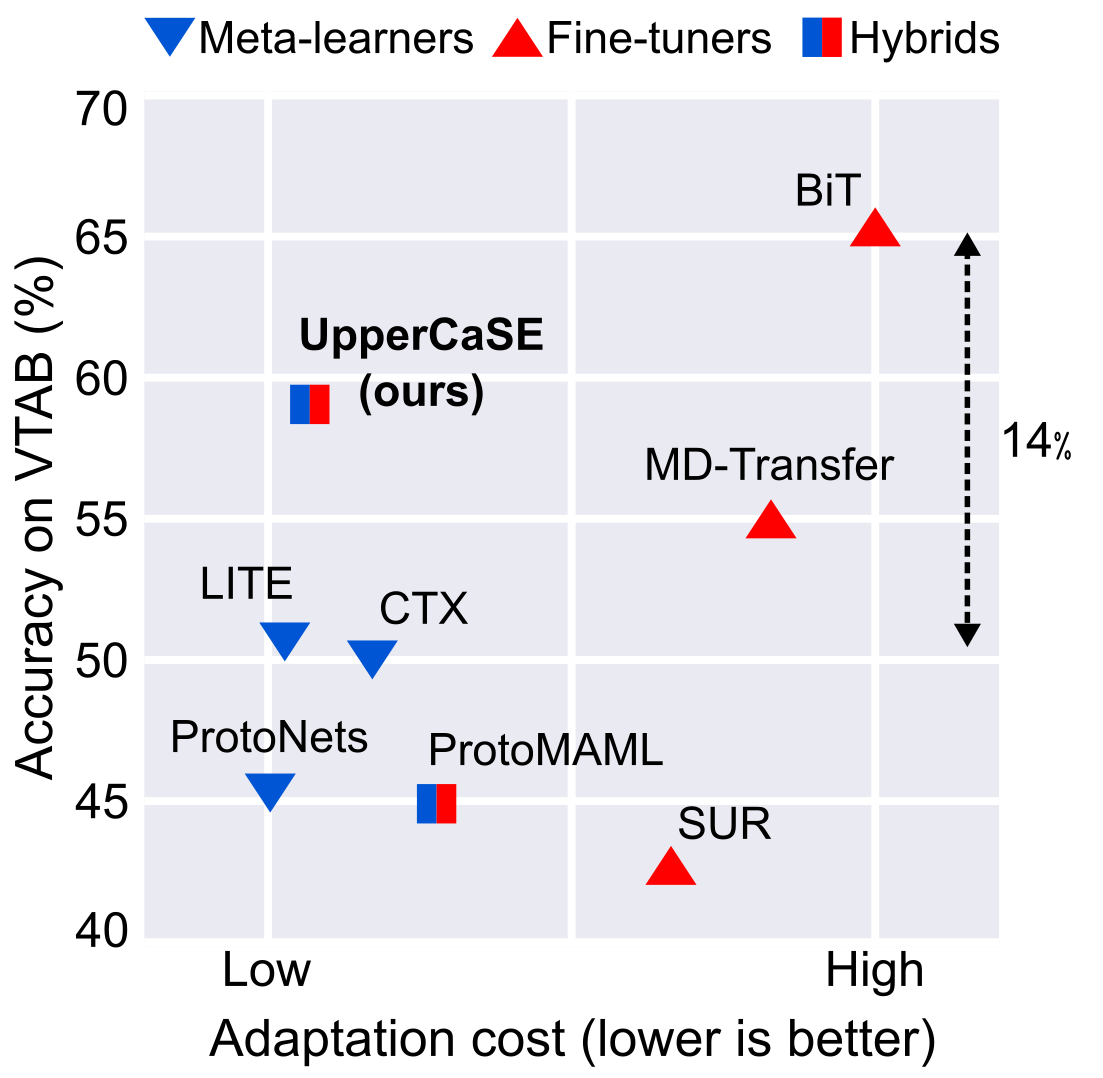}
  \vspace{-3ex}
  \caption{Accuracy and adaptation cost on VTAB for meta-learners (\textcolor{blue}{blue}), fine-tuners (\textcolor{red}{red}), and hybrids (\textcolor{blue}{blue}-\textcolor{red}{red}). Black dotted-line is the previous Pareto front across categories. UpperCaSE narrows the gap with the leading fine-tuning method and represents the best trade-off in terms of accuracy/adaptation-cost.}
  \label{fig:accuracy_vs_speed}
\end{wrapfigure}

We can gain an insight on the differences between those two paradigms by comparing them in terms of accuracy and adaptation cost. Figure~\ref{fig:accuracy_vs_speed} illustrates this comparison by showing on the vertical axis the average classification accuracy on the 18 datasets of the Visual Task Adaptation Benchmark (VTAB, \citealt{dumoulin2021comparing}), and on the horizontal axis the adaptation cost measured as the number of multiply–accumulate operations (MACs) required to adapt on a single task (see Appendix~\ref{appendix:ssec_experimental_details} for details). Overall, fine-tuners achieve a higher classification accuracy than meta-learners but are more expensive to adapt. The comparison between two state-of-the-art methods for both categories, Big Transfer (BiT, \citealt{kolesnikov2020big}) and LITE~\citep{bronskill2021memory}, shows a substantial performance gap of $14\%$ in favor of the fine-tuner but at a much higher adaptation cost, with BiT requiring $526 \times 10^{12}$ MACs and LITE only $0.2 \times 10^{12}$ MACs.

It is crucial to find solutions that retain the best of both worlds: the accuracy of fine-tuners and low adaptation cost of meta-learners. The main bottleneck that hampers the adaptation of fine-tuners is the need for multiple gradient adjustments over the entire set of network parameters. Restricting those adjustments to the last linear layer (head) significantly speeds up fine-tuning, but it harms performance (e.g.~see experiments in Section~\ref{ssec:analysis_case}). Finding a way to rapidly adapt the feature extractor (body) is therefore the main obstacle to bypass. In this paper we propose a hybrid solution to this issue, exploiting meta-learned adapters for rapidly adjusting the body and a fine-tuning routine for optimizing the head.

At the core of our approach is a novel extension of the popular Squeeze-and-Excitation block proposed by \cite{hu2018squeeze} to the meta-learning setting that we call \textbf{C}ontextu\textbf{a}l \textbf{S}queeze-and-\textbf{E}xcitation (\textbf{CaSE}). We exploit CaSE as building block of a hybrid training protocol called \textbf{UpperCaSE} which is based on the idea of adjusting the body of the network in a single forward pass over the context, and reserving the use of expensive fine-tuning routines for the linear head, similarly to methods like MetaOptNet~\citep{lee2019meta}, R2D2~\citep{bertinetto2018meta}, and ANIL~\citep{raghu2019rapid}. Figure~\ref{fig:accuracy_vs_speed} shows how UpperCaSE substantially improves the performance in the low-cost regime, outperforming meta-learners, fine-tuners such as MD-Transfer \citep{triantafillou2019meta}, and reducing the gap with the current state of the art (BiT). When adaptation cost is critical, UpperCaSE is the best method currently available since it can provide substantial computation savings and compelling classification performance. 

Our \emph{contributions} can be summarized as follows:

\begin{compactenum}
%\begin{enumerate}
    \item We introduce a new adapter called \textbf{C}ontextu\textbf{a}l \textbf{S}queeze-and-\textbf{E}xcitation (\textbf{CaSE}), based on the popular Squeeze-and-Excitation model proposed by \cite{hu2018squeeze}, that outperforms other adaptation mechanisms (e.g. the FiLM generators used in \citealt{bronskill2021memory}) in terms of parameter efficiency (a $75\%$ reduction in the number of adaptation parameters) and classification accuracy (a $1.5\%$ improvement on MetaDataset and VTAB). The code is released with an open-source license~\footnote[1]{\url{https://github.com/mpatacchiola/contextual-squeeze-and-excitation}}.
    \item We use CaSE adaptive blocks in conjuction with a fine-tuning routine for the linear head in a model called UpperCaSE, reporting an improved classification accuracy compared to the SOTA meta-learner \citep{bronskill2021memory} on the 8 datasets of MDv2 ($+2.5\%$ on average) and the 18 datasets of VTAB ($+6.8\%$ on average), narrowing the gap with BiT~\citep{kolesnikov2020big} with the benefit of orders of magnitude lower adaptation cost.
    \item We showcase the potential of UpperCaSE 
    in a real-world personalization task on the ORBIT dataset \citep{massiceti2021orbit}, where it compares favorably with the leading methods in the challenging cross-domain setting (training on MDv2, testing on ORBIT).
\end{compactenum}
%\end{enumerate}

\section{Contextual Squeeze-and-Excitation (CaSE)} \label{sec:contextual_squeeze_and_excitation}

\textbf{Problem formulation} In this paragraph we introduce the few-shot learning notation, as this will be used to describe the functioning of a CaSE adaptive block. Let us define a collection of meta-training tasks as $\mathcal{D} = \{ \tau_1, \dots, \tau_D\}$ where $\tau_i = (\mathcal{C}_i, \mathcal{T}_i )$ represents a generic task composed of a context set $\mathcal{C}_i = \{(\mathbf{x}, y)_1, \dots, (\mathbf{x}, y)_M \}$ and a target set $\mathcal{T}_i = \{(\mathbf{x}, y)_1, \dots, (\mathbf{x}, y)_D \}$ of input-output pairs. Following common practice we use the term \emph{shot} to identify the number of samples per class (e.g. 5-shot is 5 samples per class) and the term \emph{way} to identify the number of classes (e.g. 10-way is 10 classes per task). 
Given an evaluation task $\tau_{\ast} = \{\mathcal{C}_{\ast}, \mathbf{x}_{\ast} \}$ the goal is to predict the true label $y_{\ast}$ of the unlabeled target point $\mathbf{x}_{\ast}$ conditioned on the context set $\mathcal{C}_{\ast}$.

In fine-tuning methods, we are given a neural network $f_{\boldsymbol{\theta}}(\cdot)$, with parameters $\boldsymbol{\theta}$ estimated via standard supervised-learning on a large labeled dataset (e.g. ImageNet). Given a test task $\tau_{\ast}$ adaptation consists of minimizing the loss $\mathcal{L}(\cdot)$ via gradient updates to find the task-specific parameters
$\boldsymbol{\theta}_{\tau_{\ast}} \leftarrow G(\epsilon, \mathcal{L}, \tau_{\ast}, f_{\boldsymbol{\theta}})$, where $\epsilon$ is a learning rate, and $G(\cdot)$ is a functional representing an iterative routine that returns the adapted parameters $\boldsymbol{\theta}_{\tau_{\ast}}$ (used for prediction). This procedure is particularly effective because it can exploit efficient mini-batching, parallelization, and large pretrained models.

In meta-learning methods training and evaluation are performed episodically~\citep{vinyals2016matching}, with training tasks sampled from a meta-train dataset and evaluation tasks sampled from an unseen meta-test dataset. The distinction in tasks is exploited to define a hierarchy.
The parameters are divided in two groups: $\boldsymbol{\phi}$ task-common parameters shared across all tasks (top of the hierarchy), and $\boldsymbol{\psi}_\tau$ task-specific parameters estimated on the task at hand as part of an adaptive mechanism (bottom of the hierarchy). The way $\boldsymbol{\phi}$ and  $\boldsymbol{\psi}_\tau$ come into play is method dependent; they can be estimated via gradient updates (e.g. MAML, \citealt{finn2017model}), learned metrics (e.g. ProtoNets, \citealt{snell2017prototypical}), or Bayesian methods \citep{gordon2018meta, patacchiola2020bayesian, sendera2021non}.

\begin{figure}[t]
  \centering
  \includegraphics[width=1.0 \textwidth]{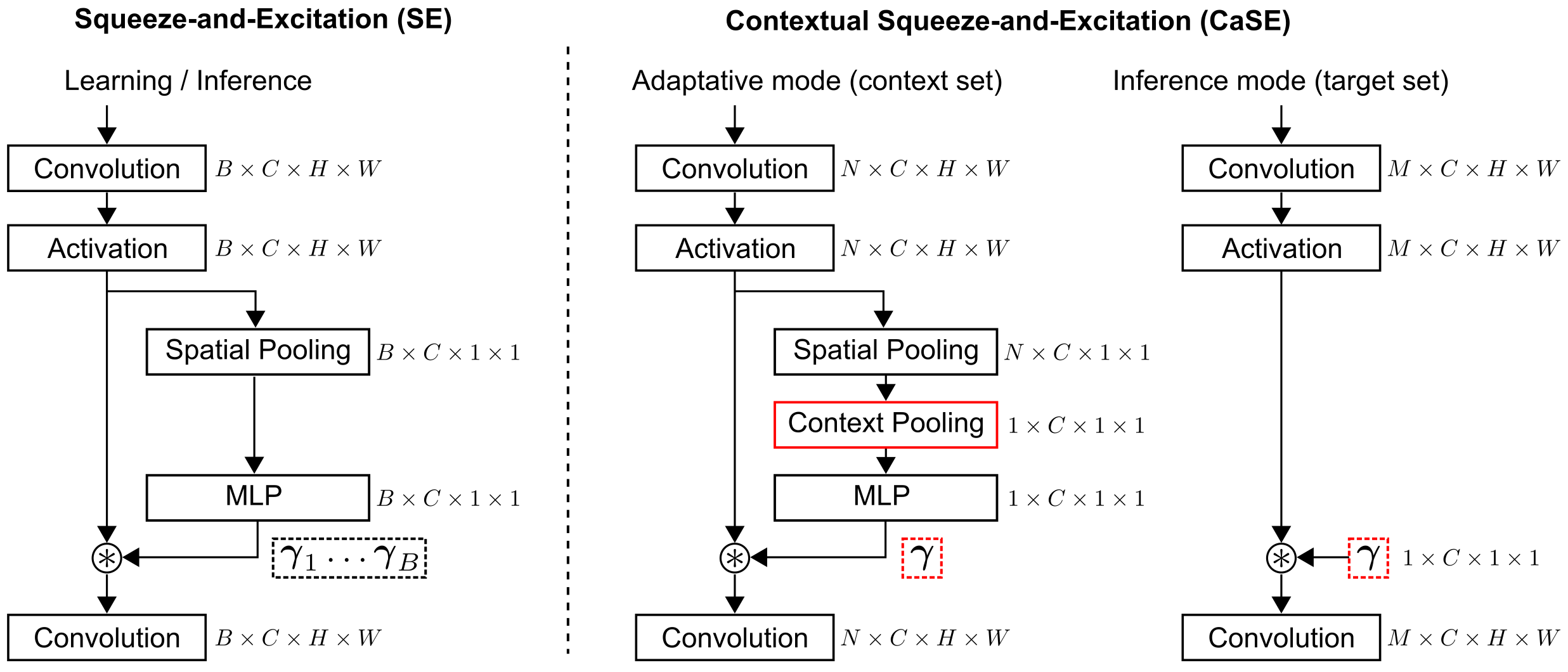}
  \caption{Comparison between the standard Squeeze-Excite (left) and the proposed Contextual Squeeze-Excite (right). \textcolor{red}{Red frames} highlight the two key differences between SE and CaSE: context pooling and scale transfer from context to target.
  $B=$ mini-batch size, $C=$ channels, $H=$ height, $W=$ width, $N=$ context-set size, $M=$ target-set size, $\ast$ elementwise multiplication.}
  \label{fig:comparison_se_case}
\end{figure}

\textbf{Standard Squeeze-Excite (SE)} We briefly introduce standard SE~\citep{hu2018squeeze}, as we are going to build on top of this work. SE is an adaptive layer used in the supervised learning setting to perform instance based channel-wise feature adaptation, which is trained following a supervised protocol together with the parameters of the neural network backbone.
Given a convolutional neural network, consider a subset of $L$ layers and associate to each one of them a Multi-Layer Perceptron (MLP), here represented as a function $g_{\boldsymbol{\phi}}(\cdot)$. The number of hidden units in the MLP is defined by the number of inputs divided by a reduction factor.
Given a mini-batch of $B$ input images, each convolution produces an output of size $B \times C \times H \times W$ where $C$ is the number of channels, $H$ the height, and $W$ the width of the resulting tensor. For simplicity we split this tensor into sub-tensors that are grouped into a set $\{ \mathbf{H}_1, \dots, \mathbf{H}_B \}$ with $ \mathbf{H}_i \in \mathbb{R}^{C \times H \times W}$. To avoid clutter, we suppress the layer indexing when possible.
SE perform a spatial pooling that produces a tensor of shape $B \times C \times 1 \times 1$; this can be interpreted as a set of vectors $\{ \mathbf{h}_1, \dots, \mathbf{h}_B \}$ with $\mathbf{h}_i \in \mathbb{R}^{C}$. For each layer $l$, the set is passed to the associated MLP that will generate an individual scale vector $\boldsymbol{\gamma}_i \in \mathbb{R}^{C}$, where
\begin{equation}
    \boldsymbol{\gamma}^{(l)}_1 = g^{(l)}_{\boldsymbol{\phi}} \left( \mathbf{h}^{(l)}_1 \right) \ \cdots \ \boldsymbol{\gamma}^{(l)}_B = g^{(l)}_{\boldsymbol{\phi}} \left( \mathbf{h}^{(l)}_B \right).
\end{equation}
An elementwise product is then performed between the scale vector and the original tensor 
\begin{equation}
    \hat{\mathbf{H}}^{(l)}_1 = \mathbf{H}_{1}^{(l)} \ast \boldsymbol{\gamma}^{(l)}_1 \ \cdots \ \hat{\mathbf{H}}^{(l)}_B = \mathbf{H}_{B}^{(l)} \ast \boldsymbol{\gamma}^{(l)}_B,
\end{equation}
with the aim of modulating the activation along the channel dimension. This operation can be interpreted as a soft attention mechanism, with the MLP conditionally deciding which channel must be attended to. A graphical representation of SE is provided in Figure~\ref{fig:comparison_se_case} (left).

\textbf{Contextual Squeeze-Excite (CaSE)} Standard SE is an instance-based mechanism that is suited for i.i.d. data in the supervised setting. In a meta-learning setting we can exploit the distinction in tasks to define a new version of SE for task-based channel-wise feature adaptation. For a task $\tau=(\mathcal{C}, \mathcal{T})$, consider the $N$ images from the context set $\mathcal{C}$, and the tensors produced by each convolution in the layers of interest $\{ \mathbf{H}_1, \dots, \mathbf{H}_N \}$ with $ \mathbf{H}_i \in \mathbb{R}^{C \times H \times W}$.
As in standard SE, we first apply a \emph{spatial} pooling to each tensor $\mathbf{H}_i$ which produces $N$ vectors $\{ \mathbf{h}_1, \dots, \mathbf{h}_N \}$ of shape $\mathbf{h}_i \in \mathbb{R}^C$. Then a \emph{context} pooling is performed; this corresponds to an empirical mean over $\{ \mathbf{h}_1, \dots, \mathbf{h}_N \}$ (see Appendix~\ref{appendix:sec_case_details} for more details about context pooling). The pooled representation is passed to the associated MLP to produce a single scale-vector for that layer
\begin{equation} \label{eq:case_pooling}
\boldsymbol{\gamma}^{(l)} = g^{(l)}_{\boldsymbol{\phi}} \left( \mathbf{\bar{h}}^{(l)} \right)
\quad \text{with} \quad 
\mathbf{\bar{h}}^{(l)} = \frac{1}{N} \left( \mathbf{h}^{(l)}_1 + \dots + \mathbf{h}^{(l)}_N \right),
\end{equation}
which is then multiplied elementwise by the original tensors
\begin{equation}
    \hat{\mathbf{H}}^{(l)}_1 = \mathbf{H}_{1}^{(l)} \ast \boldsymbol{\gamma}^{(l)} \ \cdots \ \hat{\mathbf{H}}^{(l)}_N = \mathbf{H}_{N}^{(l)} \ast \boldsymbol{\gamma}^{(l)}.
\end{equation}
The scale vector is estimated in adaptive mode and transferred to the target points $\mathcal{T}$ in inference mode (no forward pass on the MLPs), as shown in the rightmost part of Figure~\ref{fig:comparison_se_case}. In synthesis, the three major differences between SE and CaSE are: (i) CaSE uses a contextual pooling with the aim of generating an adaptive vector per-task instead of per-instance as in SE; (ii) CaSE distinguishes between an adaptive mode and an inference mode that transfers the scale from context to target, while SE does not make such a distinction; and (iii) CaSE parameters are estimated via episodic meta-training while SE parameters via standard supervised-training.
In Section~\ref{ssec:analysis_case} we show that those differences are fundamental to achieve superior performance in the few-shot setting.
A representation of a CaSE block is reported in Figure~\ref{fig:comparison_se_case}~(right), additional technical details are provided in Appendix~\ref{appendix:sec_case_details}.

\textbf{Comparison with other adapters} Popular adaptation mechanisms for few-shot learning are based on Feature-wise Linear Modulation layers (FiLM, \citealt{perez2018film}). Those mechanisms perform adaptation using a separate convolutional set-encoder to produce an embedding of the context set. The embedding is forwarded to local MLPs to produce the scale and shift vectors of the FiLM layers that modulate a pretrained model. Variations of this adapter have been used in several methods, such as TADAM~\citep{oreshkin2018tadam}, CNAPs~\citep{requeima2019fast}, SimpleCNAPs~\citep{bateni2020improved}, CAVIA~\citep{zintgraf2019fast}, and LITE~\citep{bronskill2021memory}. We will use the generic term \emph{FiLM generator} to refer to these adapters and the term \emph{FiLM} to refer to the scale and shift vectors used to modulate the activations.
There are two key differences between FiLM and CaSE: (i) CaSE exploits context pooling to aggregate the activations of the backbone instead of a separate set-encoder as in FilM generators (see Appendix~\ref{appendix:sec_case_details} for details) which is more efficient in terms of parameter count and implementation overhead; and (ii) FiLM uses scale and shift to modulate the activations, CaSE only the scale, therefore 50\% less parameters are stored in memory and transferred during inference.
In Section~\ref{ssec:analysis_case} we compare CaSE and the FiLM generators used in a recent SOTA method (LITE, \citealt{bronskill2021memory}), showing that CaSE is superior in terms of accuracy while using a fraction of the amortization parameters.

\section{UpperCaSE: system description and optimization protocol}

We exploit CaSE blocks as part of UpperCaSE, a hybrid training protocol based on Coordinate-Descent (CD). 
%that combines the benefits of meta-learning and fine-tuning. 
We call this protocol \emph{hybrid} because it combines a meta-training procedure to optimize the CaSE parameters (body) with a fine-tuning routine to estimate the task-specific parameters (head).

\textbf{Preliminaries} We are given a feature extractor (body) pretrained with supervised learning on a large dataset (e.g. ImageNet), defined as $b_{\boldsymbol{\theta}}(\cdot)$ where $\boldsymbol{\theta}$ are the pretrained parameters. CaSE blocks, parameterized by $\boldsymbol{\phi}$, are added to the model at specific locations to give $b_{\boldsymbol{\theta}, \boldsymbol{\phi}}(\cdot)$ (see Appendix~\ref{appendix:sec_case_details} for details about this step). We are interested in learning the CaSE parameters $\boldsymbol{\phi}$ keeping constant the pretrained parameters $\boldsymbol{\theta}$ (omitted from here to keep the notation uncluttered). At training time, we are given a series of tasks $\tau=\{\mathcal{C}, \mathcal{T}\} \sim \mathcal{D}$, where $\mathcal{D}$ is the training set. The number of classes (way) is calculated from the context set and used to define a linear classification head $h_{\boldsymbol{\psi}_{\tau}}(\cdot)$ parameterized by $\boldsymbol{\psi}_{\tau}$. The complete model is obtained by nesting the two functions as $h_{\boldsymbol{\psi}_{\tau}}(b_{\boldsymbol{\phi}}(\cdot))$.
We indicate a forward pass through the body over the context inputs with the shorthand $b_{\boldsymbol{\phi}}(\mathcal{C}^x) \rightarrow \{ \mathbf{z}_1, \dots, \mathbf{z}_N \}$, where $\mathbf{z}_n$ is the context embedding for the input $\mathbf{x}_n$. All the context embeddings and the associated labels are stored in $\mathcal{M} = \{(\mathbf{z}_n, y_n)\}_{n=1}^{N}$.

\textbf{Optimization challenges} We have two sets of learnable parameters, $\boldsymbol{\phi}$ the CaSE parameters, and $\boldsymbol{\psi}_\tau$ the parameters of the linear head for the task $\tau$. While $\boldsymbol{\phi}$ is shared across all tasks (task-common), $\boldsymbol{\psi}_\tau$ must be inferred on the task at hand (task-specific). In both cases, the objective is the minimization of a classification loss $\mathcal{L}$. There are some challenges in optimizing the CaSE parameters in the body, as shown by the decomposition of the full gradient
\begin{equation} \label{eq:full_gradient}
    \frac{d \mathcal{L}}{d \boldsymbol{\phi}} = \sum_\tau \left(
    \frac{\partial \mathcal{L}_\tau}{\partial \boldsymbol{\psi}_\tau} 
    \mathcolor{red}{\frac{d \boldsymbol{\psi}_\tau}{d \boldsymbol{\phi}}} +
    \frac{\partial \mathcal{L}_\tau}{\partial \boldsymbol{\phi}} \right).
\end{equation}
The first term $\partial \mathcal{L}_\tau / \partial \boldsymbol{\psi}_\tau$ (sensitivity of the loss w.r.t.~the head) and the direct gradient $\partial \mathcal{L} / \partial \boldsymbol{\phi}$ (sensitivity of the loss w.r.t.~the adaptation parameters with a fixed head) can be obtained with auto-differentiation as usual. The second term $\mathcolor{red}{d \boldsymbol{\psi}_\tau / d \boldsymbol{\phi}}$ (sensitivity of the head w.r.t.~the adaptation parameters) is problematic because $\boldsymbol{\psi}_\tau$ is obtained iteratively after a sequence of gradient updates. Backpropagating the gradients to $\boldsymbol{\phi}$ includes a backpropagation through all the gradient steps performed to obtain the task-specific $\boldsymbol{\psi}_\tau$. Previous work has showed that this produces instability, vanishing gradients, and high memory consumption \citep{antoniou2018train, rajeswaran2019meta}.

\textbf{Meta-training via Coordinate-Descent}
A potential solution to these issues is the use of implicit gradients \citep{chen2020modular, rajeswaran2019meta, chen2022meta}. The main problem with implicit gradients is the computation and inversion of the Hessian matrix as part of Cauchy’s implicit function theorem, which is infeasible when the number of parameters in the linear head is large. Another possible solution is the use of an alternating-optimization scheme, similar to the one proposed in a number of recent methods such as MetaOptNet~\citep{lee2019meta}, R2D2~\citep{bertinetto2018meta}, and ANIL~\citep{raghu2019rapid}. These methods share the idea of inner-loop-head/outer-loop-body meta-training, 
and they find the parameters of the linear head with closed form solutions or by stochastic optimization.
Starting from similar assumptions we propose a simple yet effective alternating-optimization scheme, which we formalize using Coordinate-Descent (CD) \citep{wright2015coordinate}. The idea behind CD is to consider the minimization of a complex multi-variate function as a set of simpler objectives that can be solved one at a time. In our case, we can consider the joined landscape w.r.t.~$\boldsymbol{\phi}$ and $\boldsymbol{\psi}_\tau$ as composed of two separate sets of coordinates (block CD, \citealt{wright2015coordinate}).
By minimizing $\boldsymbol{\psi}_\tau$ first, we reach a local minimum where $\partial \mathcal{L}_{\tau} / \partial \boldsymbol{\psi}_{\tau} \approx 0$. Therefore CD induces a direct optimization objective w.r.t.~$\boldsymbol{\phi}$, with Equation~\eqref{eq:full_gradient} reducing to $\partial \mathcal{L}_{\tau} / \partial \boldsymbol{\phi}$ (no \textcolor{red}{red term}).
The time complexity of this method is only affected by the number of classes but is constant w.r.t. the number of training points due to the use of mini-batching, which scales well with large tasks (e.g. those in MetaDataset and VTAB). See Appendix~\ref{appendix:sec_uppercase_details} for more details.

In practice, at each training iteration we sample a task $\tau=(\mathcal{C}, \mathcal{T}) \sim \mathcal{D}$, perform a forward pass on the body (with CaSE in adaptive mode) to get
\begin{equation} \label{eq:forward_to_embeddings}
      b_{\boldsymbol{\phi}}(\mathcal{C}^x) \rightarrow \{ \mathbf{z}_1, \dots, \mathbf{z}_N \}.
\end{equation}
The context embeddings are temporarily stored in a buffer with their associated labels $\mathcal{M} = \{ (\mathbf{z}_n, y_n) \}_{n=1}^{N}$ to avoid expensive calls to $b_{\boldsymbol{\phi}}(\cdot)$. We then set the head parameters to zero, and solve the first minimization problem (inner-loop), obtaining the task-specific parameters $\boldsymbol{\psi}_\tau$ via
\begin{equation} \label{eq:head_estimation}
   \boldsymbol{\psi}_\tau \leftarrow G\left(\alpha, \mathcal{M}, \mathcal{L}, h_{\boldsymbol{\psi}_\tau} \right)
\end{equation}
where $\alpha$ is a learning rate, and $G(\cdot)$ is a functional representing an iterative gradient-descent routine for parameter estimation (e.g. maximum likelihood estimation or maximum a posteriori estimation). Note that the iterative routine in Equation~\eqref{eq:head_estimation} only relies on the head $h_{\boldsymbol{\psi}_\tau}(\cdot)$ and not on the body $ b_{\boldsymbol{\phi}}(\cdot)$, which is the primary source of memory savings and the crucial difference with common fine-tuning methods. Moreover, the inner-loop is agnostic to the choice of optimizer, it can handle many gradient steps without complications, exploit parallelization and efficient mini-batching.

We then turn our attention to the second coordinate: the task-common parameters of the CaSE blocks in the body. For a single task, the update consists of a single optimization step w.r.t.~$\boldsymbol{\phi}$ (outer-loop) given support/target points and the task-specific parameters $\boldsymbol{\psi}_\tau$ identified previously. The final form of the equation depends on the optimizer, for a generic SGD the update is given by
\begin{equation} \label{eq:body_update}
   \boldsymbol{\phi} \leftarrow \boldsymbol{\phi} - \beta \nabla_{\boldsymbol{\phi}} \mathcal{L} \left(\mathcal{C}^y \cup Q^y, h_{\boldsymbol{\psi}_\tau}, b_{\boldsymbol{\phi}} \right),
\end{equation}
where $\beta$ is a learning rate. CaSE blocks must be in adaptive mode to allow the backpropagation of the gradients to the MLPs. The process repeats, alternating the minimization along the two sets of coordinates. The pseudo-code for train and test is provided in Appendix~\ref{appendix:sec_uppercase_details}.

\textbf{Inference on unseen tasks} After the training phase, we are given an unseen task $\tau_{\ast} = (\mathcal{C}_{\ast}, \mathbf{x}_{\ast})$ where $\mathbf{x}_{\ast}$ is a single target input and $y_{\ast}$ the associate true label to estimate. Inference consists of three steps: (i) forward pass on the body for all the context inputs with  CaSE set to adaptive mode as in Equation~\eqref{eq:forward_to_embeddings} and embeddings/labels stored in $\mathcal{M}$, (ii) estimation of the task-specific parameters $\boldsymbol{\psi}_{\ast}$ via iterative updates as in Equation~\eqref{eq:head_estimation}, and (iii) inference of the target-point membership via a forward pass over body and head $\hat{y}_{\ast} = h_{\boldsymbol{\psi}_{\ast}} \left( b_{\boldsymbol{\phi}}(\mathbf{x}_{\ast}) \right)$ with CaSE in inference mode.

\section{Related work} \label{sec:related_work}

\textbf{Meta-learning} There has been a large volume of publications related to meta-learning. Here we focus on those methods that are the most related to our work, and refer the reader to a recent survey for additional details \citep{hospedales2020meta}.
LITE~\citep{bronskill2021memory} is a protocol for training meta-learners on large images, that achieved SOTA accuracy on VTAB+MD. LITE is particularly relevant in this work, as its best performing method is based on Simple CNAPs~\citep{bateni2020improved} that exploits FiLM for fast body adaptation. We compare against LITE in Section~\ref{ssec:analysis_uppercase} showing that UpperCaSE is superior in terms of classification accuracy and parameter efficiency.

\textbf{Fine-tuning} \cite{chen2019closer} were the first to expose the potential of simple fine-tuning baselines for transfer learning. MD-Transfer has been proposed in \cite{triantafillou2019meta} as an effective fine-tuning baseline for the MetaDataset benchmark.
More recently \cite{kolesnikov2020big} have presented Big Transfer (BiT), showing that large models pretrained on ILSVRC-2012 ImageNet and the full ImageNet-21k are very effective at transfer learning. MD-Transfer and BiT differ in terms of classification head, learning schedule, normalization layers, and batching. Fine-tuning only the last linear layer can be effective \citep{bauer2017discriminative, tian2020rethinking}. We compare against this baseline in Section~\ref{ssec:analysis_case}, showing that adapting the body via CaSE significantly boosts the performance.
Overall, fine-tuners have consistently outperformed meta-learners in terms of classification accuracy, only under particular conditions (e.g. strong class-imbalance) the trend is reversed \citep{ochal2021few, ochal2021sensitive}.

%. It is only under the condition of strong class-imbalance that fine-tuners are outperformed \citep{ochal2021few, ochal2021sensitive}.

\textbf{Hybrids} Hybrid methods are trained episodically like meta-learners but rely on fine-tuning routines for adaptation. 
Model Agnostic Meta-Learning (MAML, \citealt{finn2017model}) finds a set of parameters that is a good starting point for adaptation towards new tasks in a few gradient steps. MAML has been the inspiration for a series of other models such as MAML++~\citep{antoniou2018train}, ProtoMAML~\citep{triantafillou2019meta}, and Reptile~\citep{nichol2018first}. 

\textbf{Dynamic networks} CaSE blocks belong to the wider family of dynamic networks, models that can adapt their structure or parameters to different inputs \citep{han2021dynamic}. Adaptive components have been used in a variety of applications, such as neural compression \citep{veit2018convolutional, wu2018blockdrop}, generation of artistic styles \citep{dumoulin2016learned, huang2017arbitrary}, or routing \citep{guo2019spottune}. Residual adapters \citep{rebuffi2017learning, rebuffi2018efficient} have been used in transfer learning (non few-shot) but they rely on fine-tuning routines which are significantly slow during adaptation.
More recently, \cite{li2022cross} have used serial and residual adapters in the few-shot setting, with the task-specific weights being adapted from scratch on the context set. This approach has similar limitations, since it requires backpropagation to the task-specific weights in the body of the network which is costly. In \cite{sun2019meta} the authors introduce a Meta-Transfer Learning (MTL) method for the few-shot setting. In MTL a series of scale and shift parameters are meta-learned across tasks and then dynamically adapted during the test phase via fine-tuning. This method suffers of similar limitations, as the fine-tuning stage is expensive during adaptation. Moreover, MTL relies on  scale and shift vectors to perform adaptation whereas CaSE only relies on a scale vector, meaning that it needs to store and transfer 50\% less parameters at test time.

\section{Experiments} \label{sec:experiments}

In this section we report on experiments on VTAB+MD \citep{dumoulin2021comparing} and ORBIT \citep{massiceti2021orbit}. VTAB+MD has become the standard evaluation protocol for few-shot approaches, and it includes a large number of datasets (8 test dataset for MD, 18 for VTAB).
For a description of ORBIT, see Section~\ref{ssec:analysis_uppercase}. In all experiments we used the following pretrained (on ImageNet) backbones: EfficientNetB0 from the official Torchvision repository; ResNet50x1-S released with BiT \citep{kolesnikov2020big}. We used three workstations (CPU 6 cores, 110GB of RAM, and a Tesla V100 GPU), the meta-training protocol of \cite{bronskill2021memory} ($10K$ training tasks, updates every 16 tasks), the Adam optimizer with a linearly-decayed learning rate in $[10^{-3}, 10^{-5}]$ for both the CaSE and linear-head. The head is updated 500 times using a random mini-batch of size 128. MD test results are averaged over 1200 tasks per-dataset (confidence intervals in appendix). We did not use data augmentation. Code to reproduce the experiments is available at \url{https://github.com/mpatacchiola/contextual-squeeze-and-excitation}.

\subsection{Analysis of CaSE blocks} \label{ssec:analysis_case}

In this sub-section we report empirical results related to CaSE blocks in three directions: \textbf{1)} we compared standard SE~\citep{hu2018squeeze} and CaSE on MDv2 and VTAB, confirming that a) adaptation helps over not adapting, b) contextual adaptation (CaSE) outperforms instance based adaptation (SE); \textbf{2)} we compare CaSE against a SOTA FiLM generator \citep{bronskill2021memory}, showing that CaSE is significantly more efficient using $75\%$ fewer parameters while boosting the classification accuracy on average by $+1.5\%$ on VTAB and MD2; and \textbf{3)} we provide an insight on the effectiveness of CaSE blocks with a series of qualitative analysis.

\begin{figure}[t]
  \centering
  \includegraphics[width=1.0 \textwidth]{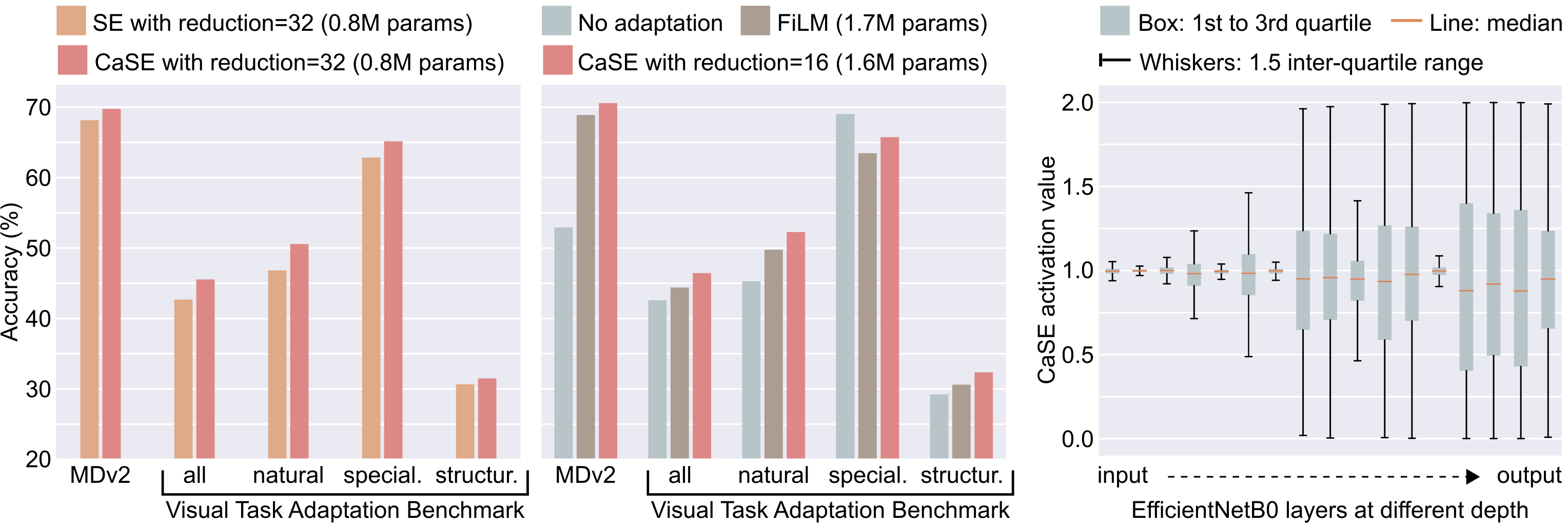}
  \caption{\textbf{Left}: CaSE vs Squeeze-and-Excitation (SE) (both methods use EfficientNetB0, $84 \times 84$ inputs, Mahalanobis-distance head). CaSE outperforms SE in all conditions. \textbf{Center}: CaSE vs.~FiLM generators \citep{bronskill2021memory} and a baseline with no body adaptation (all methods use EfficientNetB0, $84 \times 84$ inputs, Mahalanobis-distance head). CaSE outperforms FiLM generators in all conditions. \textbf{Right}: boxplot of CaSE activations at different depth of an EfficientNetB0 for 800 tasks sampled from the MDv2 test set ($224 \times 224$ inputs, UpperCaSE). The modulation of CaSE is minimal at early stages for general-purpose filters and increases at deeper stages.}
  \label{fig:barchart_case_vs_all}
\end{figure}

\textbf{Comparing SE vs.~CaSE} We compare standard SE and the proposed CaSE on VTAB and MD-v2. For a fair comparison we keep constant all factors of variation (backbone, training schedule, hyperparameters, etc.) and use the same reduction of 32 (0.8M adaptive parameters). In order to compare the results with the other experiments in this section, we use a Mahalanobis-distance head as in \cite{bronskill2021memory}, reporting results with a linear head in the appendix. We summarize the results in Figure~\ref{fig:barchart_case_vs_all}~(left) and add a tabular breakdown in Appendix~\ref{appendix:ssec_case_vs_se}. CaSE outperforms SE in all conditions, confirming that a contextual adaptation mechanism is fundamental to transfer knowledge effectively across tasks.

\textbf{Comparing adaptation mechanisms} We perform a comparison on VTAB+MD of CaSE against FiLM generators \citep{bronskill2021memory}, and a baseline that uses a pretrained model but no adaptation of the body. Methods are compared in identical conditions, using a Mahalanobis-distance head, an EfficientNetB0 backbone, and same training schedule.
We show a summary of the results in Figure~\ref{fig:barchart_case_vs_all}~(center) and provide a full breakdown in the appendix. CaSE is able to outperform FiLM generators in all conditions. In Appendix~\ref{appendix:ssec_case_vs_adapters} we report the results for CaSE with reduction 64 (0.4M parameters) showing that it is able to outperform FiLM generators (1.7M parameters) using a fraction of the parameters. The comparison with the baseline with no adaptation, shows that in all but one condition (VTAB specialized) adaptation is beneficial. This is likely due to the strong domain shift introduced by some of the specialized datasets.

\textbf{Role of CaSE blocks} To examine the role of CaSE blocks we analyze the aggregated activations at different stages of the body for 800 tasks sampled from the MDv2 test set using an EfficientNetB0 trained with UpperCaSE on $224 \times 224$ images.
In Figure~\ref{fig:barchart_case_vs_all}~(right) we report the aggregated distribution as boxplots, and in Appendix~\ref{appendix:ssec_case_role_case_blocks} we provide a per-dataset breakdown. Overall the median is close to 1.0 (identity) which is the expected behavior as on average we aim at exploiting the underlying pretrained model. The variance is small at early stages, indicating that CaSE has learned to take advantage of general-purpose filters that are useful across all tasks. In deeper layers the variance increases, showing a task-specific modulation effect. In Appendix~\ref{appendix:ssec_case_role_case_blocks} we also include a plot with per-channel activations for all datasets at different depths, showing that the modulation is similar across datasets at early stages and it diverges later on. An ablation study of different factors (e.g. reduction, number of hidden layers, activation functions) is reported in Appendix~\ref{appendix:ssec_ablation}.

\subsection{Performance evaluation of UpperCaSE} \label{ssec:analysis_uppercase}

In this sub-section we analyze the performance of UpperCaSE in two settings: \textbf{1)} comparison on the VTAB+MD benchmark against SOTA fine-tuners and meta-learners, where we show that UpperCaSE is able to outperform all the meta-learners, narrowing the gap with Big Transfer (BiT) on VTAB; \textbf{2)} we show an application of UpperCaSE in a real-world personalization task on the challenging ORBIT dataset \citep{massiceti2021orbit} for the cross-domain condition MDv2$\rightarrow$ORBIT, where we achieve the best average-score in most metrics, although these improvements are within the error bars.

\begin{table}[!h]
\caption{\textbf{UpperCaSE outperforms fine-tuners on MDv2 and narrows the gap on VTAB with the leading method (BiT) with a much lower adaptation cost}. Average accuracy on the 26 datasets of VTAB+MD. RN=ResNet, EN=EfficientNet. Img: image size. Param.: total parameters (no adapters) in millions. Cost: MACs to adapt on a task (10-shot, 100-way), in Teras. 
Best results in bold.}
\vskip 0.1in
\begin{adjustbox}{width=1.0\textwidth,center}
\begin{tabular}{lcccccccccc}
\toprule
 & & & & & \multicolumn{1}{c}{\textbf{Cost}~$\downarrow$} & \multicolumn{1}{c}{\textbf{MDv2}~$\uparrow$} & \multicolumn{4}{c}{\textbf{VTAB}~$\uparrow$} \\
 \cmidrule(lr){6-6} \cmidrule(lr){7-7} \cmidrule(lr){8-11}
 \textbf{Method} & \textbf{Protocol} & \textbf{Net} & \textbf{Img} & \textbf{Param.} & \textbf{MACs} & \textbf{all} & \textbf{all} & \textbf{natur.}  & \textbf{spec.} & \textbf{struc.} \\
\midrule
MD-Transfer & fine-tuning & RN18 & 126 & 11.2 & 118.6 & 63.4 & 55.6 & 52.4 & 72.9 & 49.3 \\ 
SUR & fine-tuning & RN50 & 224 & 164.6 & 28.8 & 71.3 & 43.7 & 50.9 & 66.2 & 27.2 \\
Big Transfer & fine-tuning & RN50 & 224 & 23.5 & 526.3 & 73.3 & \textbf{65.4} & \textbf{69.4} & \textbf{81.0} & \textbf{54.5} \\
\midrule
\emph{UpperCaSE} & hybrid & RN50 & 224 & 23.5 & 0.8 & 74.9 & 56.6 & 66.3 & 80.1 & 37.6 \\
\emph{UpperCaSE} & hybrid & ENB0 & 224 & 4.0 & \textbf{0.2} & \textbf{76.1} & 58.4 & 69.1 & 80.3 & 39.4 \\
\bottomrule
\end{tabular}
\end{adjustbox}
\label{tab:uppercase_vs_finetuners_vtab_md_results}
\end{table}

\begin{table}[!h]
\caption{\textbf{UpperCaSE outperforms all meta-learning/hybrid methods and uses the lowest number of parameters per adaptive blocks}. Average accuracy on the 26 datasets of VTAB+MD. RN=ResNet, EN=EfficientNet. Img: image size. Param.: total parameters (excluding adapters). Adapt.: total adaptive parameters in millions.
Best results in bold.}
\vskip 0.1in
\begin{adjustbox}{width=1.0\textwidth,center}
\begin{tabular}{lcccccccccc}
\toprule
 & & & & & \multicolumn{1}{c}{\textbf{Adapt.}~$\downarrow$} & \multicolumn{1}{c}{\textbf{MDv2}~$\uparrow$} & \multicolumn{4}{c}{\textbf{VTAB}~$\uparrow$} \\
 \cmidrule(lr){6-6} \cmidrule(lr){7-7} \cmidrule(lr){8-11}
 \textbf{Method} & \textbf{Protocol} & \textbf{Net} & \textbf{Img} & \textbf{Param.} & \textbf{count} & \textbf{all} & \textbf{all} & \textbf{natur.} & \textbf{spec.} & \textbf{struc.} \\
\midrule
ProtoMAML & hybrid & RN18 & 126 & 11.2 & n/a & 64.2 & 45.0 & 45.7 & 70.7 & 31.5 \\
CTX & meta-learning & RN34 & 224 & 21.3 & n/a & 71.6 & 50.5 & 61.1 & 67.3 & 34.0 \\
%ProtoNet & meta-learning & RN18 & 11.2 & 60.3 & 40.2 & 44.0 & 57.6 & 28.7 \\
ProtoNet & meta-learning & ENB0 & 224 & 4.0 & n/a & 72.7 & 46.1 & 60.9 & 64.2 & 25.9 \\
LITE & meta-learning & ENB0 & 224 & 4.0 & 1.7 & 73.8 & 51.4 & 65.2 & 71.9 & 30.8 \\
\midrule
\emph{UpperCaSE} & hybrid & RN50 & 224 & 23.5 & 0.8 & 74.9 & 56.6 & 66.3 & 80.1 & 37.6 \\
\emph{UpperCaSE} & hybrid & ENB0 & 224 & 4.0 & \textbf{0.4} & \textbf{76.1} & \textbf{58.4} & \textbf{69.1} & \textbf{80.3} & \textbf{39.4} \\
\bottomrule
\end{tabular}
\end{adjustbox}
\label{tab:uppercase_vs_metalearners_vtab_md_results}
\end{table}

\textbf{Comparison on VTAB+MD} We compare UpperCaSE against fine-tuners, meta-learners, and hybrids on the 18 datasets of VTAB and the 8 datasets of MetaDataset-v2 (MDv2) and report the results in Table~\ref{tab:uppercase_vs_finetuners_vtab_md_results} and Table~\ref{tab:uppercase_vs_metalearners_vtab_md_results}. UpperCaSE outperforms all methods (including BiT) on MDv2 with an accuracy of $74.9\%$ (ResNet50) and $76.1\%$ (EfficientNetB0). On VTAB, UpperCaSE outperforms most methods, narrowing the gap with BiT. A closer look at the differences in performance on VTAB between UpperCaSE and BiT (see Table~\ref{tab:uppercase_vs_finetuners_vtab_md_results}) shows that the gap is narrower on the natural and specialized splits ($+3.1\%$ and $+0.9\%$) but larger on structured ($+16.9\%$).

The breakdown by dataset reported in Appendix~\ref{appendix:ssec_uppercsae_results_vtab_md} shows that the major performance drops are on tasks that require localization and counting (e.g.~dSprites, SmallNORB). Similar issues are encountered by methods such as LITE~\citep{bronskill2021memory} which are based on FiLM generators, suggesting that those tasks may introduce a strong domain shift w.r.t.~the meta-training set that is difficult to compensate without fine-tuning the body. It is not clear whether transfer learning is beneficial on these datasets in the first place. The results in terms of adaptation cost (see Table~\ref{tab:uppercase_vs_finetuners_vtab_md_results}) over a synthetic task (10-shot, 100 way) show that UpperCaSE is orders of magnitude more efficient ($0.2 \times 10^{12}$ MACs) than all fine-tuners, with BiT being the most expensive method overall ($526.3 \times 10^{12}$ MACs). The comparison against meta-learners in terms of number of adaptive parameters (see Table~\ref{tab:uppercase_vs_metalearners_vtab_md_results}) shows that UpperCaSE requires a fraction of the parameters (0.4 vs 1.7 millions for an EfficientNetB0) compared to LITE~\citep{bronskill2021memory} which is based on FiLM generators.

\textbf{Comparison on ORBIT} We compare UpperCaSE to other methods on ORBIT \citep{massiceti2021orbit}, a real-world dataset for teachable object recognizers. ORBIT consists of 3822 videos of 486 objects recorded by 77 blind/low-vision people on their mobile phones. The dataset is challenging because objects are poorly framed, occluded, blurred, and in a wide variation of backgrounds and lighting. The dataset includes two sets of target videos, one for clean video evaluation (CLE-VE) with well-centered objects, and another for clutter video evaluation (CLU-VE) with objects in complex, cluttered environments. We consider a hard transfer-learning condition where classifiers are meta-trained on MetaDataset and tested on ORBIT. 

Results are reported in Table~\ref{tab:orbit_results}. UpperCaSE outperforms all other methods (on average) on most metrics, being within error bars with the two leading methods. Comparing UpperCaSE with FineTuner, the gap in favor of UpperCaSE is marginal on CLU-VE but substantial on CLE-VE (frame accuracy $+2\%$, video accuracy $+1.8\%$, and FTR $-2.7$). Comparison in terms of adaptation cost (average MACs over all tasks) shows that UpperCaSE is orders of magnitude more efficient than FineTuner and close to the leading method (ProtoNet).

\begin{table}[t]
\caption{\textbf{ORBIT: UpperCaSE obtains the best average-score in most metrics, being within error bars with leading methods.}  Average accuracy and 95\% confidence interval for frames, videos, and frames to recognition (FTR). Cost: average MACs over all tasks (Teras). Results and setup from \cite{massiceti2021orbit}: meta-train on MetaDataset and test on ORBIT, image-size $84 \times 84$, ResNet18 backbone, 85 test tasks (17 test users, 5 tasks per user). Best results (within error bars) in bold.}
\vskip 0.1in
\begin{adjustbox}{width=1.0\textwidth,center}
\begin{tabular}{lccccccc}
\toprule
 & \multicolumn{1}{c}{\textbf{Cost}} & \multicolumn{3}{c}{\textbf{Clean Video Evaluation (CLE-VE)}} & \multicolumn{3}{c}{\textbf{Clutter Video Evaluation (CLU-VE)}} \\
 \cmidrule(lr){2-2} \cmidrule(lr){3-5} \cmidrule(lr){6-8}
 \textbf{Method} & \textbf{MACs}$\downarrow$ & \textbf{frame acc.}$\uparrow$ & \textbf{FTR}$\downarrow$ & \textbf{video acc.}$\uparrow$ & \textbf{frame acc.}$\uparrow$ & \textbf{FTR}$\downarrow$  & \textbf{video acc.}$\uparrow$\\
\midrule
ProtoNet & \textbf{3.2} & \textbf{59.0$\pm$2.2} & \textbf{11.5$\pm$1.8} & \textbf{69.2$\pm$3.0} & \textbf{47.0$\pm$1.8} & \textbf{20.4$\pm$1.7} & \textbf{52.8$\pm$2.5}\\
CNAPs & 3.5 & 51.9$\pm$2.5 & 20.8$\pm$2.3 & 60.8$\pm$3.2 & 41.6$\pm$1.9 & 30.7$\pm$2.1 & 43.0$\pm$2.5\\
MAML & 95.3  & 42.5$\pm$2.7 & 37.3$\pm$3.0 & 47.0$\pm$3.2 & 24.3$\pm$1.8 & 62.3$\pm$2.3 & 25.7$\pm$2.2 \\
FineTuner & 317.7 & \textbf{61.0$\pm$2.2} & \textbf{11.5$\pm$1.8} & \textbf{72.6$\pm$2.9} & \textbf{48.4$\pm$1.9} & \textbf{19.1$\pm$1.7} & \textbf{54.1$\pm$2.5} \\
\midrule
\emph{UpperCaSE} & 3.5 & \textbf{63.0$\pm$2.2} & \textbf{8.8$\pm$1.6} & \textbf{74.4$\pm$2.8} & \textbf{48.1$\pm$1.8} & \textbf{18.2$\pm$1.7} & \textbf{54.5$\pm$2.5}\\
\bottomrule
\end{tabular}
\end{adjustbox}
\label{tab:orbit_results}
\end{table}

\section{Conclusions} \label{sec:conclusions}

We have introduced a new adaptive block called CaSE, which is based on the popular Squeeze-and-Excitation (SE) block proposed by \cite{hu2018squeeze}. CaSE is effective at modulating a pretrained model in the few-shot setting, outperforming other adaptation mechanisms. Exploiting CaSE we have designed UpperCaSE, a hybrid method based on a Coordinate-Descent training protocol, that combines the performance of fine-tuners with the low adaptation cost of meta-learners. UpperCaSE achieves SOTA accuracy w.r.t.~meta-learners on the 26 datasets of VTAB+MD and it compares favorably with leading methods in the ORBIT personalization benchmark. 

\textbf{Limitations} There are two \emph{limitations} that are worth mentioning: (i) UpperCaSE requires iterative gradient updates that are hardware-dependent and may be slow/unavailable in some portable devices; (ii) breakdown VTAB results per-dataset shows that the method falls short on structured datasets. This indicates that fine-tuning the body may be necessary for high accuracy when the shift w.r.t.~the meta-training set is large.

\textbf{Societal impact} Applications based on CaSE and UpperCaSE could be deployed in few-shot classification settings that can have a positive impact such as: medical diagnosis, recommendation systems, object detection, etc. The efficiency of our method can reduce energy consumption and benefit the environment. Certain applications require careful consideration to avoid biases that can harm specific groups of people (e.g. surveillance, legal decision-making).

\begin{ack}
Funding in direct support of this work: 
Massimiliano Patacchiola, John Bronskill, Aliaksandra Shysheya, and Richard E. Turner are supported by an EPSRC Prosperity Partnership EP/T005386/1 between the EPSRC,
Microsoft Research and the University of Cambridge. The authors would like to thank: anonymous reviewers for useful comments and suggestions; Aristeidis Panos, Daniela Massiceti, and Shoaib Ahmed Siddiqui for providing suggestions and feedback on the preliminary version of the manuscript.
\end{ack}

%\section*{References}
\bibliography{neurips_2022}
\bibliographystyle{apalike}

%%%%%%%%%%%%%%%%%%%%%%%%%%%%%%%%%%%%%%%%%%%%%%%%%%%%%%%%%%%%

%%%%%%%%%%%%%%%%%%%%%%%%%%%%%%%%%%%%%%%%%%%%%%%%%%%%%%%%%%%%

\clearpage
\appendix

%--------------------
\section{CaSE: additional details} \label{appendix:sec_case_details}

\subsection{CaSE implementation} \label{appendix:ssec_techincal_details_case}

\textbf{Standardization} Empirically we have observed that standardizing the pooled representations before passing them to the MLP improves the training stability in CaSE (but not in SE). Standardization is performed by taking the pooled representation at layer $l$ as showed in Equation~\eqref{eq:case_pooling}, that is $\mathbf{\bar{h}}^{(l)} \in \mathbb{R}^{C}$, subtracting the mean and dividing by the standard deviation. 

\textbf{Activation function for the output layer} Standard SE blocks usually rely on a sigmoid function in the last layer of the MLPs. This works well when the adaptive block is trained in parallel with the underlying neural network. However, in our case we use a pretrained model and learning can be speeded up considerably by enforcing the identity function as output of the MLPs. We achieve this by multiplying the output of the sigmoid by a constant scalar $c=2$ which extends the range to $[0, 2]$, and then set to zero the weights and bias of the layer. This has the effect of enforcing the identity function at the beginning of the training. We have also used a linear activation function instead of a sigmoid, with good results. When using a linear output the identity can be enforced by setting the weights of the last layer to zero, and the bias to one.
An ablation over the activation function of SE and CaSE is provided in Appendix~\ref{appendix:ssec_ablation} (Table~\ref{tab:ablation_activation_functions}).

\textbf{CaSE location} For the choice of CaSE location in the feature extractor, we followed the same principles used in \cite{bronskill2021memory} for FiLM generators. In EfficientNetB0 we place CaSE at the beginning of each hyperblock and the last layer (excluding the first layer). Differently from FiLM (placed after the BatchNorm) we place CaSE after the non-linearity (as done in standard SE) and before the Squeeze-and-Excitation block (included by default in EfficientNet):

\begin{center}
Conv2d$\rightarrow$BatchNorm2d$\rightarrow$SiLU$\rightarrow$CaSE$\rightarrow$SqueezeExcitation$\rightarrow$Conv2d$\rightarrow$BatchNorm2d
\end{center}

This results in a total of 18 CaSE blocks for EfficientNetB0. Increasing the number of blocks did not provide a significant benefit. In ResNet18 we place two CaSE blocks per each basic block as: 

\begin{center}
Conv2d$\rightarrow$BatchNorm2d$\rightarrow$ReLU$\rightarrow$CaSE$\rightarrow$Conv2d$\rightarrow$BatchNorm2d$\rightarrow$ReLU$\rightarrow$CaSE
\end{center}

Similarly we place two CaSE blocks inside a bottleneck block in ResNet50. See the code for more details. 

Based on the qualitative analysis reported in Section~\ref{sec:experiments} we hypothesize that adaptive blocks are not needed in the initial layers of the network, since at those stages their activity is minimal. Identifying which layer needs adapters and which layer does not, can reduce even more the parameter count of adaptive blocks. Additional work is needed to fully understand this factor.

\textbf{CaSE reduction} The number of parameters allocated to the CaSE blocks is regulated by a divider $r$ that is used to compute the number of hidden units in the MLPs. Given the input size $C$ (corresponding to the number of channels in that layer) the number of hidden units is given by $C / r$.
We also use a clipping factor $r_{min}$ that prevents the number of units to fall under a given threshold. This prevents the allocation of a low number of units for layers with a small number of channels.

\subsection{Context pooling}

In this section we provide additional details about the context pooling operation performed in a CaSE adaptive block (described in Section~\ref{sec:contextual_squeeze_and_excitation}). 

\textbf{Similarities with other methods} Context pooling is a way to summarize a task with a permutation-invariant aggregation of the embeddings. A similar mechanism has been exploited in various meta-learning methods. For instance, in ProtoNets~\citep{snell2017prototypical} a prototype for a single class is computed by taking the average over all the context embeddings associated to the inputs for that class. The embeddings are generated in the last layer of the feature extractor. In Simple-CNAPs~\citep{bateni2020improved} a prototype is estimated as in ProtoNets but it is used to define a Gaussian distribution instead of a mean vector. Neural latent variable models, such as those derived from the Neural Processes family \citep{garnelo2018neural} also rely on similar permutation-invariant aggregations to define
distributions over functions.

\textbf{Global vs. local context-pooling} Comparing CaSE with the FiLM generators of \cite{bronskill2021memory} it is possible to distinguish between two types of context pooling: global and local. The FiLM generators of \cite{bronskill2021memory} rely on a \emph{global} pooling strategy, meaning that the aggregation is performed once-for-all by using a dedicated convolutional set encoder. More specifically, the encoder takes as input all the context images and produces embeddings for each one of them, followed by an average-pooling of those embeddings. The aggregated embedding is then passed to MLPs in each layer that generates a scale and shift parameter. Crucially, each MLP receives the same embedding.

CaSE exploits a \emph{local} context-pooling at the layer level. The convolutional set encoder is discarded, and the feature maps produces by the backbone itself at each stage are used as context embeddings. Therefore, the MLPs responsible for generating the scale parameters receive a unique embedding. As showed in the experimental section (Section~\ref{sec:experiments}), local pooling improves performances and uses less parameters, as no convolutional encoder is needed. Additional details about the differences between CaSE and FiLM generators is also provided in the paper (Section~\ref{sec:related_work}).

\clearpage
\subsection{Pytorch code for CaSE}

Implementation of a CaSE adaptive block in Pytorch. The script is also available as \texttt{case.py} at \url{https://github.com/mpatacchiola/contextual-squeeze-and-excitation}.

\inputpython{./code/case.py}{1}{100}

%--------------------
\section{UppereCaSE: additional details} \label{appendix:sec_uppercase_details}

\subsection{Algorithm of UpperCaSE} \label{appendix:ssec_algorithm}

\begin{algorithm}[H]
\small
\caption{UpperCaSE: training function for the few-shot classification setting.}
\label{alg:train_overview}
\textbf{Require:}  $\mathcal{D} = \{\tau_1, \dots, \tau_D\}$ training dataset \\
\textbf{Require:} $b_{\boldsymbol{\phi}}()$ pretrained feature extractor (body) with CaSE blocks parameterized by $\boldsymbol{\phi}$. \\
\textbf{Require:} \texttt{step()}: gradient-step function; $\mathcal{L}$ loss; $\alpha$, $\beta$: step-size hyperparameters for the optimizer.
\begin{algorithmic}[1]
\vspace{0.1cm} %empty line
\State Set $\boldsymbol{\phi}$ to random values \Comment{optional: set $\boldsymbol{\phi}$ to enforce identity in CaSE output}
\While{not done}
    \State Sample task $\tau=(\mathcal{C}, \mathcal{T}) \sim \mathcal{D}$
    \State Forward pass over context set $b_{\boldsymbol{\phi}}(\mathcal{C}^x) \ \rightarrow \ \mathbf{z}_1, \dots, \mathbf{z}_N$ \Comment{CaSE in adaptive mode}
    \State Store context embeddings and associated labels $\mathcal{M} = \{(\mathbf{z}_n, y_n)\}_{n=1}^{N}$ \Comment{temporary memory buffer}
    \State Define a linear model for the head $h_{\boldsymbol{\psi}_{\tau}}()$ and set $\boldsymbol{\psi}_{\tau}$ to zero
    \For{\texttt{total inner-steps}} \Comment{loop to estimate head params}
        \State Sample (with replacement) mini-batch of training pairs $\mathcal{B} \sim \mathcal{M}$
        \State Update the head parameters $\boldsymbol{\psi}_{\tau} \leftarrow$ \texttt{step($\alpha, \mathcal{L}, \mathcal{B}, h_{\boldsymbol{\psi}_{\tau}}$)}
    \EndFor
    \State Update the CaSE parameters $\boldsymbol{\phi} \leftarrow$ \texttt{step($\beta, \mathcal{L}, \mathcal{C}, \mathcal{T}, b_{\boldsymbol{\phi}}, h_{\boldsymbol{\psi}_{\tau}}$)} \Comment{CaSE in adaptive mode}
\EndWhile
\vspace{0.1cm} %\item[] %empty line
\Statex
\end{algorithmic}
  \vspace{-0.4cm}%
\end{algorithm}

\begin{algorithm}[H]
\small
\caption{UpperCaSE: test function for the few-shot classification setting.}
\label{alg:test_overview}
\textbf{Require:}  $\tau_{\ast} = ( \mathcal{C}_{\ast}, \mathbf{x}_{\ast} )$ unseen test task with target input $\mathbf{x}_{\ast}$ an context $\mathcal{C}_{\ast}$. \\
\textbf{Require:} $b_{\boldsymbol{\phi}}()$ pretrained feature extractor (body) with meta-learned CaSE blocks parameterized by $\boldsymbol{\phi}$. \\
\textbf{Require:} \texttt{step()}: gradient-step function; $\mathcal{L}$ loss;  $\alpha$: step-size hyperparameter for the optimizer.
\begin{algorithmic}[1]
\vspace{0.1cm}
    \State Forward pass over context set $b_{\boldsymbol{\phi}}(\mathcal{C}^x_{\ast}) \ \rightarrow \ \mathbf{z}_1, \dots, \mathbf{z}_N$ \Comment{CaSE in adaptive mode}
    \State Store context embeddings and associated labels $\mathcal{M}_{\ast} = \{(\mathbf{z}_n, y_n)\}_{n=1}^{N}$ \Comment{temporary memory buffer}
    \State Define a linear model for the head $h_{\boldsymbol{\psi}_{\tau_{\ast}}}()$ and set $\boldsymbol{\psi}_{\tau_{\ast}}$ to zero
    \For{\texttt{total inner-steps}} \Comment{loop to estimate head params}
        \State Sample (with replacement) mini-batch of training pairs $\mathcal{B}_{\ast} \sim \mathcal{M}_{\ast}$
        \State Update the head parameters $\boldsymbol{\psi}_{\tau_{\ast}} \leftarrow$ \texttt{step($\alpha, \mathcal{L}, \mathcal{B}_{\ast}, h_{\boldsymbol{\psi}_{\tau_{\ast}}}$)}
    \EndFor
    \State \textbf{Return} Prediction $\hat{y}_{\ast} = h_{\boldsymbol{\psi}_{\tau_{\ast}}}(b_{\boldsymbol{\phi}}(\mathbf{x}_{\ast}))$ \Comment{CaSE in inference mode}
\vspace{0.1cm}
\Statex
\end{algorithmic}
  \vspace{-0.4cm}%
\end{algorithm}

%--------------------
\section{Additional experimental details and results} \label{appendix:sec_additional_experimental_results}

\subsection{Additional details} \label{appendix:ssec_experimental_details}

\textbf{MACs counting} MACs are proportional to the size of the task, size of the images, and number of classes. We can count MACs using synthetic tasks. In our case we used a synthetic task of 100-way, 10-shot with input images of size $224 \times 224 \times 3$ generated via Gaussian noise ($\mu=0, \sigma=1$), and labels generated as random integers. We used a mini-batch of size 128 and 500 update steps for UpperCaSE and BiT with an EfficientNetB0 backbone for the first and a ResNet50-S for the second. For MD-Transfer we used the same parameters reported in \cite{dumoulin2021comparing} with images of size $126 \times 126 \times 3$ and ResNet18 backbone. For the ORBIT experiments we counted MACs by using the code in the original repository \footnote{\url{https://github.com/microsoft/ORBIT-Dataset}} and reporting the average MACs over all test tasks for both CLE-VE and CLU-VE using a ResNet18 backbone.

\textbf{VTAB+MD training} We follow the protocol reported in the original papers \citep{triantafillou2019meta, dumoulin2021comparing} training UpperCaSE for 10K tasks on the training datasets and evaluating on the MD test set and on the VTAB datasets. At evaluation time we sample 1200 tasks from the MD test set, and report the mean and confidence intervals. On VTAB we report the results of a single run on the test data (data points are given in advance and do not change across seeds).
In all experiments we used the MetaDataset-v2 (MDv2) which does not include ImageNet in the test set. We used a pretrained EfficientNetB0 from the official Torchvision repository \footnote{\url{https://pytorch.org/vision}}, and a pretrained ResNet50-S from the BiT repository \footnote{\url{https://github.com/google-research/big_transfer}}. We normalized the inputs using the values reported in the Torchvision documentation (mean=[0.485, 0.456, 0.406], std=[0.229, 0.224, 0.225]), for ResNet50-S we use the BiT normalization values (mean=[0.5, 0.5, 0.5], std=[0.5, 0.5, 0.5]).

\textbf{ORBIT training} For the ORBIT experiments we trained UpperCaSE on MDv2 using a pretrained ResNet18 taken from the official Torchvision repository. We normalized the inputs using the values reported in the Torchvision documentation (mean=[0.485, 0.456, 0.406], std=[0.229, 0.224, 0.225]).
For the evaluation phase we followed the instructions reported in \cite{massiceti2021orbit}. 

\subsection{CaSE vs SE} \label{appendix:ssec_case_vs_se}

\begin{table}[H]
\caption{Comparing CaSE against standard Squeeze-and-Excitation (SE) on VTAB+MD using different adaptation heads. MD: Mahalanobis distance head \citep{bronskill2021memory}. Linear: linear head trained with UpperCaSE.  All adaptive blocks use a reduction of 32. Best results in bold.}
\vskip 0.15in
\begin{center}
\begin{tabular}{lcc|cc}
\toprule
Model & SE & CaSE & SE & CaSE \\
Contextual pooling & No & Yes & No & Yes \\
Adaptation head & MD & MD & Linear & Linear \\
Image size & 84 & 84 & 224 & 224\\
\midrule
MetaDataset (all) & 67.8 & \textbf{69.6} & 74.6 & \textbf{76.2} \\
VTAB (all) & 43.6 & \textbf{45.3} & 56.6 & \textbf{58.2} \\
VTAB (natural) & 47.5 & \textbf{50.2} & 65.3 & \textbf{68.1} \\
VTAB (specialized) & 63.6 & \textbf{64.9} & \textbf{79.8} & 79.6 \\
VTAB (structured) & 30.6 & \textbf{31.8} & 38.6 & \textbf{40.1} \\
\bottomrule
\end{tabular}
\label{tab:se_vs_case}
\end{center}
\vskip -0.1in
\end{table}

\subsection{CaSE vs other adapters} \label{appendix:ssec_case_vs_adapters}

\begin{table}[H]
\caption{Comparing CaSE adaptive blocks (with reduction 64, 32, 16) on VTAB+MD against the FiLM generators used in \cite{bronskill2021memory}, and a baseline with no body adaptation. CaSE blocks are more efficient in terms of adaptive and amortization parameters while providing higher classification accuracy. All models have been trained and tested on $84 \times 84$ images, using a Mahalanobis distance head. Best results in bold.}
\vskip 0.15in
\begin{center}
\begin{tabular}{lccccc}
\toprule
Adaptation type & None & FiLM & CaSE64 & CaSE32 & CaSE16 \\
Adaptive Params (M) & n/a & 0.02 & \textbf{0.01} & \textbf{0.01} & \textbf{0.01} \\
Amortiz. Params (M) & n/a & 1.7 & \textbf{0.4} & 0.8 & 1.6\\
\midrule
MetaDataset (all) & 53.4 & 68.4 & 69.8 & 69.6 & \textbf{70.4}\\
VTAB (all) & 43.5 & 44.7 & 46.2 & 45.3 & \textbf{46.4}\\
VTAB (natural) & 45.4 & 49.5 & 52.1 & 50.2 & \textbf{52.6}\\
VTAB (specialized) & \textbf{69.4} & 63.8 & 66.3 & 64.9 & 65.5\\
VTAB (structured) & 29.1 & 31.7 & 31.8 & 31.8 & \textbf{32.1}\\
\bottomrule
\end{tabular}
\end{center}
\vskip -0.1in
\end{table}

\subsection{Ablation studies} \label{appendix:ssec_ablation}

In this section we provide additional experimental results focusing on ablation studies of the CaSE adaptive block. The results can be summarized as follows:

\begin{itemize}
    \item Ablation of the activation function for the output layer for both CaSE and SE. We have tested three activation funcitons: linear, sigmoid, sigmoid with multiplier. The sigmoid with multiplier uses a constant value set to 2 to center the sigmoid at 1 (this enforces the identity function). The empirical results reported in Table~\ref{tab:ablation_activation_functions} show that the sigmoid with multiplier and the linear layer provide the best results.
    \item Ablation of the number of hidden units in the hidden layers of CaSE. The number of hidden units is controlled by the reduction and min-units parameters in the code and it depends on the number of inputs. See the paper for more details. The results reported in Table~\ref{tab:ablation_hidden_units} show that blocks with more units provide marginal gains or no gains at all. This is probably due to overfitting issues affecting the models with more units.
    \item Ablation of the number of hidden layers of CaSE. The results reported in Table~\ref{tab:ablation_hidden_layers} show that the best performance is obtained with 1 and 2 layers. The performance worsen when there are 3 or more layers which is likely due to overfitting issues affecting the models with more parameters.
    \item Ablation of the activation function for the hidden layers. Results reported in Table~\ref{tab:ablation_hidden_activation} show that CaSE is quite robust against this factor when activations like ReLU and SiLU are used but the performance worsen with Tanh. We have chosen SiLU for the experiments as this is the same activation typically used in Squeeze-and-Excitation layers (e.g. in EfficientNet backbones).
\end{itemize}

\begin{table}[H]
\caption{Performance on VTAB+MD for various activation functions used in the last layer of SE and CaSE. Sigmoid-2 indicates that the output of a standard Sigmoid is multiplied by 2. Both SE and CaSE use a reduction factor of 32 with min-clipping of 32. All model have been trained using an EfficientNetB0 backbone with a linear head on images of size $224 \times 224$. Results for SE with linear activation have not been reported because the training was unstable (loss rapidly diverging at the first iterations). Best results in bold.}
\vskip 0.15in
\begin{center}
\begin{tabular}{lccccc}
\toprule
Adaptive block & SE & SE & CaSE & CaSE & CaSE \\
Activation (output) & Sigmoid & Sigmoid-2 & Linear & Sigmoid & Sigmoid-2 \\
\midrule
MetaDataset (all) & 74.2 & 74.6 & 75.8 & 74.9 & \textbf{76.2}\\
VTAB (all) & 56.8 & 56.6 & \textbf{58.4} & 56.8 & 58.2\\
VTAB (natural) & 67.0 & 65.3 & \textbf{68.3} & 67.1 & 68.1\\
VTAB (specialized) & \textbf{81.1} & 79.8 & 79.5 & 80.8 & 79.6\\
VTAB (structured) & 36.9 & 38.6 & \textbf{40.3} & 37.1 & 40.1\\
\bottomrule
\end{tabular}
\label{tab:ablation_activation_functions}
\end{center}
\vskip -0.1in
\end{table}

\begin{table}[H]
\caption{Comparing CaSE adaptive blocks with different number of hidden layers on VTAB+MD. All models have been trained and tested on $224 \times 224$ images, using CaSE with reduction 64 and clip factor (min-units) 16, using UpperCaSE and EfficientNetB0 backbone. Best results in bold.}
\vskip 0.15in
\begin{center}
\begin{tabular}{lccccccc}
\toprule
\# Hidden layers & 1 & 2 & 3 & 4\\
Amortiz. Params (M) & \textbf{0.420} & 0.426 & 0.432 & 0.438\\
\midrule
MetaDataset (all) & 76.0 & \textbf{76.1} & 75.5  & 75.2\\
VTAB (all) & 58.2 & \textbf{58.4} & 58.2 & 58.0\\
VTAB (natural) & 68.3 & \textbf{69.1} & 68.0  & 67.4\\
VTAB (specialized) & 79.7 & 80.3 & \textbf{80.5}  & 80.3 \\
VTAB (structured) & \textbf{40.0} & 39.4 & 39.7  & 39.7\\
\bottomrule
\end{tabular}
\label{tab:ablation_hidden_layers}
\end{center}
\vskip -0.1in
\end{table}

\begin{table}[H]
\caption{Comparing CaSE adaptive blocks with different number of hidden units on VTAB+MD. The number of hidden units depends on the input size and is defined by the reduction and the clip factor (min-units). All models have been trained and tested on $224 \times 224$ images, using UpperCaSE and EfficientNetB0 backbone. Best results in bold.}
\vskip 0.15in
\begin{center}
\begin{tabular}{lcccccc}
\toprule
Reduction factor & 64 & 32 & 16 & 8\\
Clip factor & 16 & 32 & 48 & 64\\
Amortiz. Params (M) & \textbf{0.4} & 0.8 & 1.6 & 3.0\\
\midrule
MetaDataset (all) & 76.1 & \textbf{76.2} & 75.8 & \textbf{76.2}\\
VTAB (all) & 58.4 & 58.2 & 57.9 & \textbf{58.5}\\
VTAB (natural) & \textbf{69.1} & 68.1 & 67.9 & 68.3\\
VTAB (specialized) & \textbf{80.3} & 79.6 & 79.4 & 79.0\\
VTAB (structured) & 39.4 & 40.1 & 39.7 & \textbf{40.9}\\
\bottomrule
\end{tabular}
\label{tab:ablation_hidden_units}
\end{center}
\vskip -0.1in
\end{table}

\begin{table}[H]
\caption{Comparing CaSE adaptive blocks with different activation functions for the hidden layers on VTAB+MD. All models are based on a reduction factor of 64 and a clip factor of 16 (0.4M amortization parameters) and they have been trained and tested on $224 \times 224$ images, using UpperCaSE and EfficientNetB0 backbone. Best results in bold.}
\vskip 0.15in
\begin{center}
\begin{tabular}{lcccccc}
\toprule
Activation (hidden) & SiLU & ReLU & Tanh\\
\midrule
MetaDataset (all) & \textbf{76.1} & 75.8 & 74.8\\
VTAB (all) & \textbf{58.4} & 57.8 & 48.2\\
VTAB (natural) & 69.1 & \textbf{69.8} & 67.0\\
VTAB (specialized) & \textbf{80.3} & 79.7 & 80.8\\
VTAB (structured) & \textbf{39.4} & \textbf{39.4} & 36.4\\
\bottomrule
\end{tabular}
\label{tab:ablation_hidden_activation}
\end{center}
\vskip -0.1in
\end{table}

\clearpage
\subsection{Role of CaSE blocks} \label{appendix:ssec_case_role_case_blocks}

\begin{figure}[H]
  \centering
  \includegraphics[width=1.0 \textwidth]{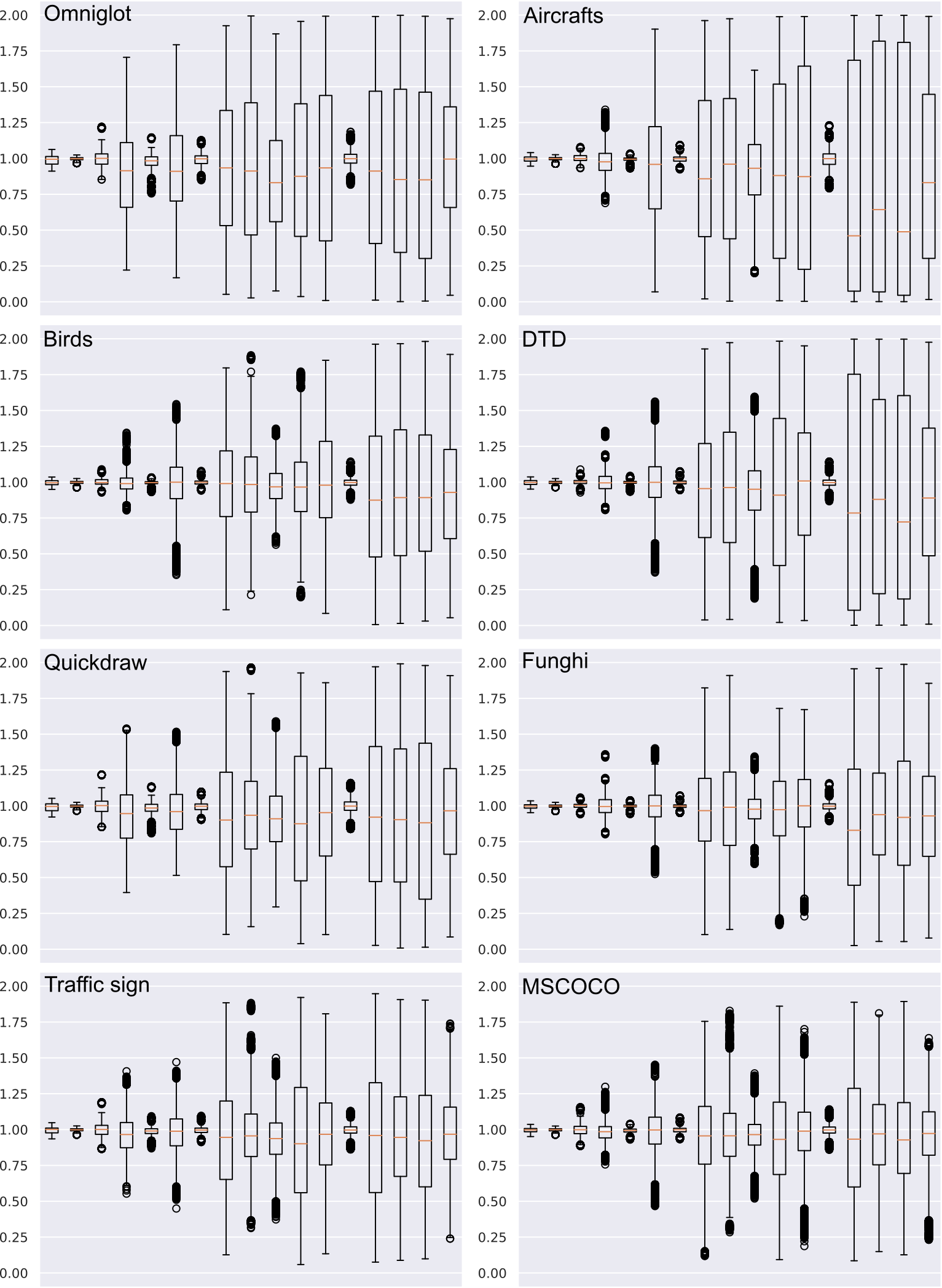}
  \caption{Boxplots for all the MDv2 test datasets (100 tasks per dataset) reporting the CaSE activation (vertical axis) at different stages of an EfficientNetB0 (horizontal axis, with early stages on the left). The box encloses first to third quartile, with the median represented by the orange line. The whiskers extend from the box by 1.5 the inter-quartile range. Outlier (point past the end of the whiskers) are represented with black circles.}
  \label{fig:appendix_boxplot_perdataset}
\end{figure}

\begin{figure}[H]
  \centering
  \includegraphics[width=0.78 \textwidth]{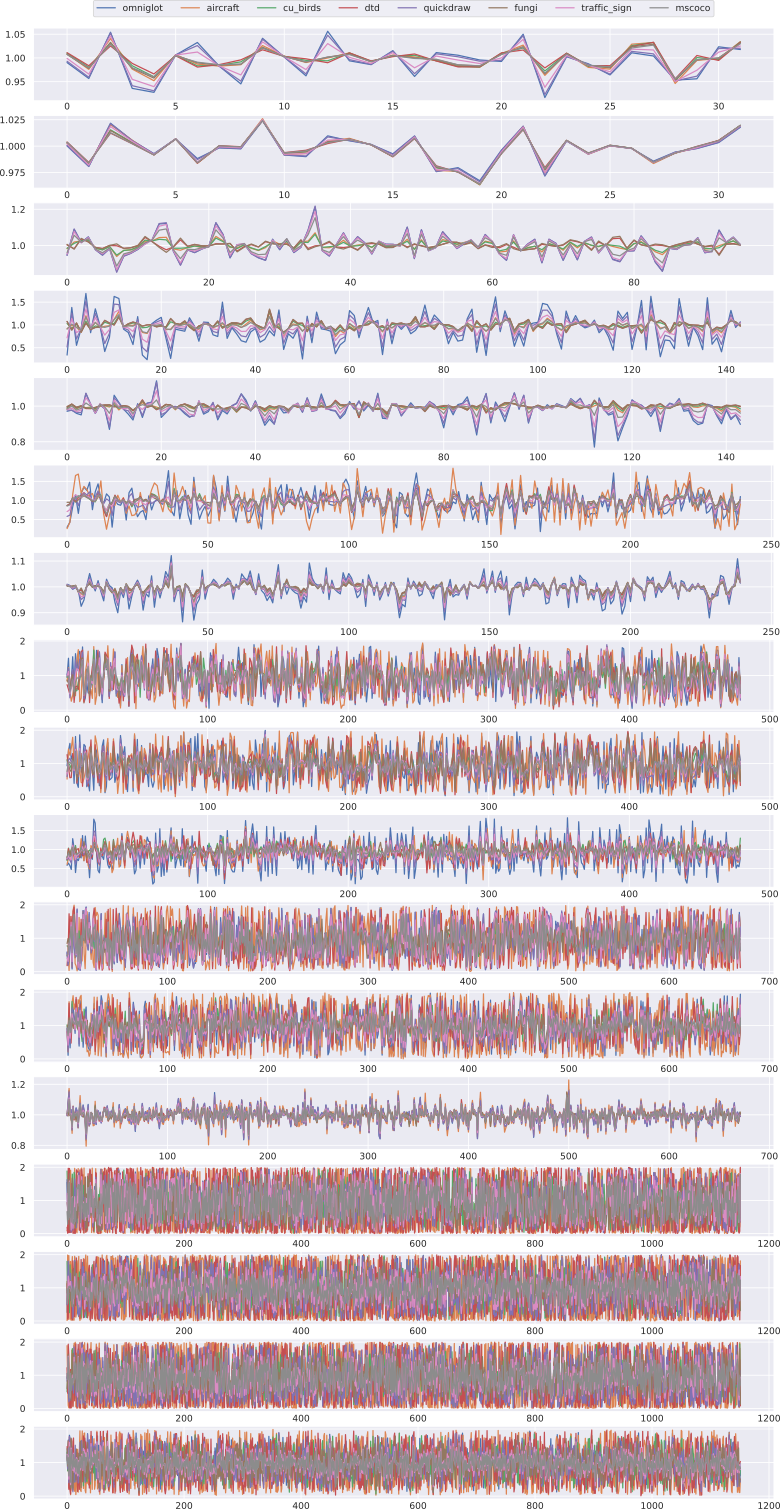}
  \caption{CaSE activation values (vertical axis) for all channels (horizontal axis) at different stages (top plots are early stages) in EfficientNetB0 for the MDv2 test dataset (one task per dataset). Values are similar and closer to one in the first stages but diverge in the latest. The magnitude tends to increase with depth.}
  \label{fig_gammas_per_layer}
\end{figure}

\clearpage

\subsection{UpperCaSE: results on VTAB+MD} \label{appendix:ssec_uppercsae_results_vtab_md}

In this section we provide a full breakdown of the results for UpperCaSE vs. other methods on the VTAB+MD benchmark. Results for other methods are taken from \cite{bronskill2021memory} and \cite{dumoulin2021comparing}. UpperCaSE uses CaSE with reduction 64 (min-clip 16) for EfficientNetB0 and reduction 32 (min-clip 32) for ResNet50-S. Results for UpperCaSE on MD are the average over 1200 test tasks. 

In Table~\ref{tab:breakdown_uppercase_vs_finetuning} we report the results for UpperCaSE against fine-tuning methods (BiT, MD-Trasnfer, SUR) and in Table~\ref{tab:breakdown_uppercase_vs_metalearning} the results for UpperCaSE against meta-learning and hybrid methods (ProtoNet, ProtoMAML, Cross Transformer CTX, LITE). Overall UpperCaSE performs well on MD and the natural split of VTAB, this may be due to the fact that transfer learning is more beneficial on those datasets as they are more similar to those used during meta-training. The largest difference in performance between UpperCaSE and fine-tuning methods is on the structured split of VTAB, which includes tasks that require counting and pose estimation. This is likely due to the difference w.r.t. the meta-training set. In this case, fine-tuning the entire network is more effective than body adaptation as the knowledge gap is wider and it requires more adjustments to the parameters.

\begin{table}[H]
\caption{Comparing UpperCaSE against fine-tuning methods. Best result in bold.}
\vskip 0.15in
\begin{center}
\begin{tabular}{lccccc}
\toprule
Model & BiT & MD-Transfer & SUR & UpperCaSE & UpperCaSE\\
%Protocol & Tune & FineTune & FiLM-64 & ATTA-64\\
Image Size & 224 & 126 & 224 & 224 & 224\\
Network & RN50-S & RN18 & RN50$\times$7 & ENB0 & RN50-S\\
Params (M) & 23.5 & 11.2 & 164.5 & 4.0 & 23.5 \\
\midrule
Omniglot & 68.0$\pm$4.5 & 82.0$\pm$1.3 & \textbf{92.8$\pm$0.5} & 90.7$\pm$0.4 & 89.1$\pm$0.5\\
Aircraft & 77.4$\pm$3.5 & 76.8$\pm$1.2 & 84.4$\pm$0.6 & \textbf{89.4$\pm$0.4} & 87.5$\pm$0.4\\
Birds & \textbf{90.8$\pm$1.5} & 61.2$\pm$1.3 & 75.8$\pm$1.0 & \textbf{90.4$\pm$0.4} & \textbf{89.6$\pm$0.4}\\
DTD & \textbf{85.0$\pm$2.5} & 66.0$\pm$1.1 & 74.3$\pm$0.7 & \textbf{83.4$\pm$0.4} & \textbf{84.8$\pm$0.5}\\
QuickDraw & 66.6$\pm$3.7 & 61.3$\pm$1.1 & 70.3$\pm$0.7 & \textbf{76.8$\pm$0.5} & 73.7$\pm$0.6\\
Fungi & 59.4$\pm$4.2 & 35.5$\pm$1.1 & \textbf{81.7$\pm$0.6} & 59.3$\pm$0.8 & 56.8$\pm$0.8\\
Traffic Sign & 73.5$\pm$4.7 & \textbf{84.7$\pm$0.9} & 50.0$\pm$1.1 & 68.5$\pm$0.8 & 70.6$\pm$0.8\\
MSCOCO & \textbf{65.7$\pm$2.7} & 39.6$\pm$1.0 & 49.4$\pm$1.1 & 50.8$\pm$0.7 & 46.7$\pm$0.8\\
\midrule
Caltech101 & 87.2 & 70.6 & 82.3 & \textbf{88.3} & 86.2\\
CIFAR100 & \textbf{54.4} & 31.3 & 33.7 & 52.7 & 47.0\\
Flowers102 & 83.3 & 66.1 & 55.7 & \textbf{85.3} & 83.0\\
Pets & 87.9 & 49.1 & 76.3 & 89.9 & 89.3\\
Sun397 & 33.3 & 13.9 & 27.5 & \textbf{35.8} & 32.5\\
SVHN & \textbf{70.4} & 83.2 & 18.7 & 62.7 & 59.8\\
\midrule
EuroSAT & \textbf{94.4} & 88.7 & 78.9 & 92.2 & 91.6\\
Resics45 & \textbf{76.1} & 63.7 & 62.4 & 75.5 & 74.4\\
Patch Camelyon & \textbf{83.1} & 81.5 & 75.6 & 79.3 & 80.9\\
Retinopathy & 70.2 & 57.6 & 27.9 &  \textbf{74.3} & 73.7\\
\midrule
CLEVR-count & \textbf{74.0} & 40.3 & 30.0 & 40.3 & 42.0\\
CLEVR-dist & 51.5 & \textbf{52.9} & 37.1 & 38.9 & 37.3\\
dSprites-loc & 82.7 & \textbf{85.9} & 30.0 & 45.3 & 38.1\\
dSprites-ori & \textbf{55.1} & 46.4 & 19.8 & 42.5 & 41.4\\
SmallNORB-azi & 17.8 & \textbf{36.5} & 12.9 & 15.7 & 15.1\\
SmallNORB-elev & \textbf{32.1} & 31.2 & 18.1 & 22.7 & 21.0\\
DMLab & \textbf{43.2} & 37.9 & 33.3 & 38.7 & 36.1\\
KITTI-dist & \textbf{79.9} & 58.7 & 52.3 & 71.0 & 69.6\\
\midrule
MetaDataset (all) & 73.3 & 63.4 & 71.0 & \textbf{76.1} & 74.9\\
VTAB (all) & \textbf{65.4} & 55.6 & 42.9 & 58.4 & 56.6\\
VTAB (natural) & \textbf{69.4} & 52.4 & 49.0 & 69.1 & 66.3\\
VTAB (specialized) & \textbf{81.0} & 72.9 & 61.2 & 80.3 & 80.1\\
VTAB (structured) & \textbf{54.5} & 49.4 & 29.2 & 39.4 & 37.6\\
\bottomrule
\end{tabular}
\label{tab:breakdown_uppercase_vs_finetuning}
\end{center}
\vskip -0.1in
\end{table}

\begin{table}[H]
\caption{Comparing UpperCaSE against meta-learning and hybrid methods. Best result in bold.}
\vskip 0.15in
\begin{center}
\begin{small}
\begin{tabular}{lcccccc}
\toprule
Model & ProtoNet & ProtoMAML & CTX & LITE & UpperCaSE & UpperCaSE \\
%Protocol & Tune & FineTune & FiLM-64 & ATTA-64\\
Image Size & 224 & 126 & 224 & 224 & 224 & 224\\
Network & ENB0 & RN18 & RN34 & ENB0 & ENB0 & RN50-S \\
Params (M) & 4.0 & 11.2 & 21.3 & 4.0 & 4.0 & 23.5 \\
\midrule
Omniglot & 88.3$\pm$0.8 & \textbf{90.2$\pm$0.7} & 84.6$\pm$0.9 & 86.5$\pm$0.8 & \textbf{90.7$\pm$0.4} & 89.1$\pm$0.5\\
Aircraft & 85.0$\pm$0.7 & 82.1$\pm$0.6 & 85.3$\pm$0.8 & 83.6$\pm$0.7 & \textbf{89.4$\pm$0.4} & 87.5$\pm$0.4\\
Birds & \textbf{90.2$\pm$0.5} & 73.4$\pm$0.9 & 72.9$\pm$1.1 & 88.6$\pm$0.7 & \textbf{90.4$\pm$0.4} & 89.6$\pm$0.4\\
DTD & 81.4$\pm$0.6 & 66.3$\pm$0.8 & 77.3$\pm$0.7 & \textbf{84.1$\pm$0.7} & 83.4$\pm$0.4 & \textbf{84.8$\pm$0.5}\\
QuickDraw & \textbf{76.0$\pm$0.7} & 66.4$\pm$1.0 & 73.3$\pm$0.8 & \textbf{75.7$\pm$0.8} & 59.3$\pm$0.8 & 56.8$\pm$0.8\\
Fungi & 57.4$\pm$1.1 & 46.3$\pm$1.1 & 48.0$\pm$1.2 & 56.9$\pm$1.2 & \textbf{59.3$\pm$0.8} & 56.8$\pm$0.8\\
Traffic Sign & 53.5$\pm$1.1 & 50.3$\pm$1.1 & \textbf{80.1$\pm$1.0} & 65.8$\pm$1.1 & 68.5$\pm$0.8 & 70.6$\pm$0.8\\
MSCOCO & \textbf{49.8$\pm$1.1} & 39.0$\pm$1.0 & \textbf{51.4$\pm$1.1} & \textbf{50.0$\pm$1.0} & \textbf{50.8$\pm$0.7} & 46.7$\pm$0.8\\
\midrule
Caltech101 & 87.4 & 73.1 & 84.2 & 87.7 & \textbf{88.3} & 86.2\\
CIFAR100 & 43.1 & 29.7 & 37.5 & 48.8 & \textbf{52.7} & 47.0\\
Flowers102 & 78.2 & 60.2 & 81.8 & 83.5 & \textbf{85.3} & 83.0\\
Pets & 88.6 & 56.6 & 70.9 & 89.3 & \textbf{89.9} & 89.3\\
Sun397 & 32.9 & 8.1 & 24.8 & 30.9 & \textbf{35.8} & 32.5\\
SVHN & 35.2 & 46.8 & \textbf{67.2} & 51.0 & 62.7 & 59.8\\
\midrule
EuroSAT & 83.3 & 80.1 & 86.4 & 89.3 & \textbf{92.2} & 91.6\\
Resics45 & 68.8 & 53.5 & 67.7 & \textbf{76.4} & 75.5 & 74.4\\
Patch Camelyon & 73.3 & 75.9 & 79.8 & \textbf{81.4} & 79.3 & 80.9\\
Retinopathy & 31.3 & 73.2 & 35.5 & 40.3 & \textbf{74.3} & 73.7\\
\midrule
CLEVR-count & 27.2 & 32.7 & 27.9 & 31.4 & 40.3 & \textbf{42.0}\\
CLEVR-dist & 28.5 & 35.4 & 29.6 & 32.8 & \textbf{38.9} & 37.3\\
dSprites-loc & 13.4 & 42.0 & 23.2 & 12.3 & \textbf{45.3} & 38.1\\
dSprites-ori & 19.6 & 23.0 & \textbf{46.9} & 31.1 & 42.5 & 41.4\\
SmallNORB-azi & 9.4 & 13.4 & \textbf{37.0} & 14.5 & 15.7 & 15.1\\
SmallNORB-elev & 17.0 & 18.8 & 21.6 & 21.0 & \textbf{22.7} & 21.0\\
DMLab & 35.8 & 32.5 & 31.9 & \textbf{39.4} & 38.7 & 36.1\\
KITTI-dist & 56.5 & 54.4 & 54.3 & 63.9 & \textbf{71.0} & 69.6\\
\midrule
MetaDataset (all) & 72.7 & 64.2 & 71.6 & 73.9 & \textbf{76.1} & 74.9\\
VTAB (all) & 46.1 & 45.0 & 50.5 & 51.4 & \textbf{58.4} & 56.6\\
VTAB (natural) & 60.9 & 45.7 & 61.1 & 65.2 & \textbf{69.1} & 66.3\\
VTAB (specialized) & 64.2 & 70.7 & 67.3 & 71.9 & \textbf{80.3} & 80.1\\
VTAB (structured) & 25.9 & 31.5 & 34.1 & 30.8 & \textbf{39.4} & 37.6\\
\bottomrule
\end{tabular}
\label{tab:breakdown_uppercase_vs_metalearning}
\end{small}
\end{center}
\vskip -0.1in
\end{table}

\end{document}